\documentclass[12pt]{article}
\usepackage{amssymb}                         
\usepackage[T2A]{fontenc}
\usepackage{enumerate,amsmath,graphicx}
\usepackage{graphicx}

\newtheorem{theorem}{\indent Theorem}
\newtheorem{lemma}{\indent Lemma}
\newtheorem{corollary}{\indent Corollary}
\newtheorem{proposition}{\indent Proposition}
\newtheorem{definition}{\indent Definition}

\newcommand{\since}{since }
\newcommand {\x}{\mbox{\bf x}}
\newcommand {\y}{\mbox{\bf y}}
\newcommand {\A}{\mbox{\bf A}}
\newcommand {\e}{\mbox{\bf e}}
\newcommand {\q}{\mbox{\bf q}}
\newcommand {\p}{\mbox{\bf p}}
\newcommand {\z}{\mbox{\bf z}}
\newcommand{\ie}{i.\,e. }
\def\R{\mathbb R}

\begin{document}

\pagestyle{plain}
\title{Optimal Recombination in Genetic Algorithms}

\author{Anton V. Eremeev, Julia V. Kovalenko }

\maketitle
\begin{center}
Sobolev Institute of Mathematics, \\
Laboratory of Discrete Optimization,\\
630090, Novosibirsk, Russia.\\
Email:~eremeev@ofim.oscsbras.ru\\
Omsk F.M.~Dostoevsky State University,\\
Institute of Mathematics and Information Technologies,\\
644077, Omsk, Russia.\\
Email:~juliakoval86@mail.ru
\end{center}

\begin{abstract}
This paper surveys results on complexity of the optimal
recombination problem~(ORP), which consists in finding the best
possible offspring as a result of a recombination operator in a
genetic algorithm, given two parent solutions. We consider
efficient reductions of the ORPs, allowing to establish polynomial
solvability or NP-hardness of the ORPs, as well as direct proofs
of hardness results.\\

2) Optimal Recombination\\
3) Complexity\\
4) Crossover
\end{abstract}

\section{Introduction \label{intro}}

The genetic algorithms~(GAs) originally suggested by
J.~Holland~\cite{Holl75} are randomized heuristic search methods
using an evolving population of sample solutions, based on analogy
with the genetic mechanisms in nature. Various modifications of
GAs have been widely used in operations research, pattern
recognition, artificial intelligence, and other areas (see
e.g.~\cite{Reev,YI96,Yank99}). Despite numerous experimental
studies of these algorithms, the theoretical analysis of their
efficiency is currently at an early stage~\cite{BSW}. Efficiency
of GAs depends significantly on the choice of {\em crossover}
operator, that combines the given {\em parent} solutions, aiming
to produce "good"\ {\em offspring} solutions (see
e.g.~\cite{JW02}). Originally the crossover operator was proposed
as a simple randomized procedure~\cite{Holl75}, but subsequently
the more elaborated problem-specific crossover operators
emerged~\cite{Reev}.

This paper is devoted to complexity and solution methods of the
{\em Optimal Recombination Problem}~(ORP), which consists in
finding the best possible offspring as a result of a crossover
operator, given two feasible parent solutions. The ORP is a
supplementary problem (usually) of smaller dimension than the
original problem, formulated in view of the basic principles of
crossover~\cite{R94}.

The first GAs using the optimal recombination appeared in the
works of C.C.~Agarwal, J.B.~Orlin and R.P.~Tai~\cite{AOT} and
M.~Yagiura and T.~Ibaraki~\cite{YI96}. These works provide GAs for
the Maximum Independent Set problem and several permutation
problems. Subsequent results in~\cite{BN98,CT,DEG10,ErKo11,GLM00}
and other works added more experimental support to expediency of
solving the optimal recombination problems in crossover operators.

Interestingly, it turned out that a number of NP-hard optimization
problems have efficiently solvable ORPs. The present paper
contains a survey of results focused on the issue of efficient
solvability vs. intractability of the ORPs.



The paper is structured as follows. The formal definition of the
ORP for NP optimization problems is introduced in
Section~\ref{sec:OptRecDef}. Then, using efficient reductions
between the ORPs it is shown in Section~\ref{sec:PolySolve} that
the optimal recombination is computable in polynomial time for the
Maximum Weight Set Packing Problem, the Minimum Weight Set
Partition Problem and for one of the versions of the Simple Plant
Location Problem. In Section~\ref{sec:PolySolve} we also propose
an efficient optimal recombination operator for the Boolean Linear
Programming Problems with at most two variables per inequality. In
Section~\ref{sec:NPHard} we consider a number of NP-hard ORPs for
the Boolean Linear Programming Problems. The computational
complexity of ORP for the Travelling Salesman Problem is
considered in Section~\ref{sec:TSP} both for the symmetric and for
the general case. Strong NP-hardness of these optimal
recombination problems is proven and solving approaches are
proposed. A closely related problem of Makespan Minimization on
Single Machine is considered in Section~\ref{sec:makespan}: it is
shown that on one hand this ORP problem is strongly NP-hard, on
the other hand, almost all of its instances are efficiently
solvable. Section~\ref{Conclusion} is devoted to the concluding
remarks and issues for further research.


\section{Optimal Recombination in Genetic Algorithms\label{sec:OptRecDef}}

We will employ the standard definition of an NP~optimization
problem (see e.g.~\cite{ACGKMP}). By $\{0,1\}^*$ we denote the set
of all strings with symbols from~$\{0,1\}$ and arbitrary string
length. For a string~$S\in \{0,1\}^*$, the symbol~$|S|$ will
denote its length. The term {\em polynomial time} stands for the
computation time which is upper bounded by a polynomial in length
of the input data. Let ${\R}_+$ denote the set of non-negative
reals.

\begin{definition}\label{def:NPO} An NP~optimization problem $\Pi$
is a triple
${\Pi=(\mbox{\rm Inst},\mbox{\rm Sol},f_I)}$, where $\mbox{\rm
Inst} \subseteq \{0,1\}^*$
is the set of instances of~$\Pi$ and:\\

1. The relation $\mbox{\rm Inst}$ is computable in polynomial time.\\

2. Given an instance $I \in \mbox{\rm Inst}$, $\mbox{\rm
Sol}(I)\subseteq \{0,1\}^{n(I)}$ is the set of feasible solutions
of~$I$, where~$n(I)$ stands for the dimension of the space of
solutions. Given $I\in \mbox{\rm Inst}$ and $\x\in
\{0,1\}^{n(I)}$, the decision whether $\x\in \mbox{\rm Sol}(I)$
may be done in polynomial time, and $n(I) \le \mbox{\rm
poly}(|I|)$ for some polynomial~$\mbox{\rm
poly}$.\\

3. Given an instance~$I \in \mbox{\rm Inst}$,  $f_I: \mbox{\rm
Sol}(I) \to {\R}_+$ is the objective function (computable in
polynomial time) to be maximized if $\ \Pi$ is an NP~maximization
problem or to be minimized if $\ \Pi$ is an NP~minimization
problem.
\end{definition}

For the sake of compactness of notation we will simply put ${\rm
Sol}$ instead of ${\rm Sol}(I)$, $n$ instead of $n(I)$ and $f$
instead of $f_I$, when it is clear what problem instance is
implied.

Throughout the paper we use the term {\em efficient algorithm} as
a synonym for polynomial-time algorithm. A problem which is solved
by such an algorithm is {\em polynomially solvable.}

Often it is possible to formulate an NP~optimization problem as a
Boolean Linear Programming Problem:
\begin{equation}\label{goal}
\mbox{max} \ \    f(\x) = \sum_{j=1}^{n} c_{j}x_{j},
\end{equation}
subject to
\begin{equation} \label{ineq}
\sum_{j=1}^{n} a_{ij} x_{j} \leq b_i, \quad i=1,\dots,m,
\end{equation}
\begin{equation} \label{bools}
x_j \in \{0,1\}, \quad j=1,\dots,n.
\end{equation}
In the context of Boolean Linear Programming Problem, $\x\in
\{0,1\}^n$ is treated as a column vector of Boolean variables
$x_1,\dots,x_n$, which belongs to~${\rm Sol}$ iff the
constraints~(\ref{ineq}) are satisfied. The similar problems where
instead of~"$\le$"\ in~(\ref{ineq}) stands~"$\ge$" or "$=$" for
some indices~$i$ (or for all~$i$) can be easily transformed to
formulation~(\ref{goal})--(\ref{bools}).
The minimization problems can be considered, using the goal
function with coefficients~$c_j$ of opposite sign. Where
appropriate, we will use a more compact notation for problem
(\ref{goal})--(\ref{bools}):
$$
\max\left\{{\bf cx} : {\bf Ax} \le {\bf b}, \ {\bf x} \in
\{0,1\}^n\right\},
$$
where~$\A$ is an ($m\times n$)-matrix with elements~$a_{ij}$,
${\bf b}=(b_1,\dots,b_m)^T$ and ${\bf c}=(c_1,\dots,c_n)$.

\subsection{Genetic Algorithms}\label{subsec:GA}

The simple GA proposed in~\cite{Holl75} has been intensively
studied and exploited over four decades (see e.g.~\cite{RR02}).
This algorithm operates with populations~$X^t, \ t=1,2,\dots$ of
binary strings in~$\{0,1\}^n$ traditionally called {\it
genotypes}. Each population consists of a fixed number of
genotypes~$N$, which is assumed to be even. In a {\em selection
operator}~${\rm Sel}$, each parent is drawn from the previous
population~$X^t$ independently with probability distribution
assigning each genotype a probability proportional to its {\em
fitness}, where fitness is measured by the value of the objective
function or a composition of the objective function with some
monotonic function.

\begin{figure}
\begin{center}
\includegraphics[height=5.16cm,width=15cm]{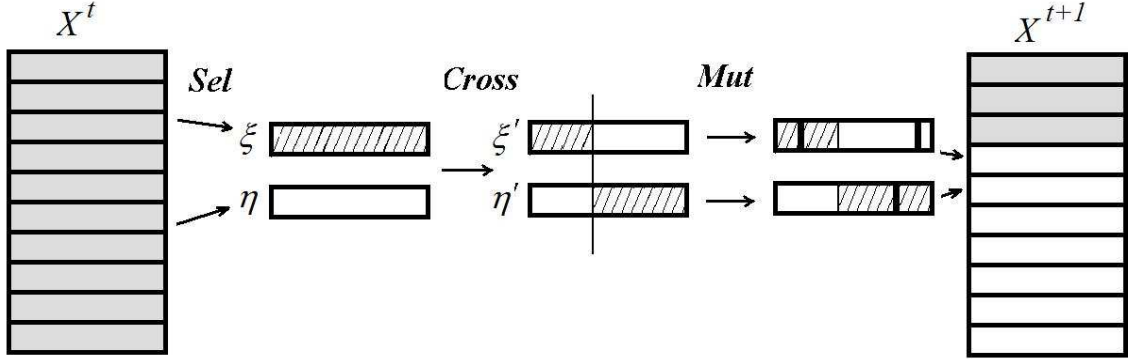}
\end{center}
\vspace{-1em} \caption{Selection, crossover and mutation in Simple
Genetic Algorithm.}\label{fig:SGA}
\end{figure}

A pair of offspring genotypes is created through recombination and
mutation stages (see Fig~\ref{fig:SGA}). In the recombination
stage, a crossover operator~${\rm Cross}$
exchanges random substrings between pairs of parent
genotypes~$\xi,\eta$ with a given constant probability~$P_{\rm c}$
so that
$$
{\bf P}\left\{\xi'=(\xi_{1},...,\xi_{j},\eta_{j+1},...,\eta_{n}),
\
\eta'=(\eta_{1},...,\eta_{j},\xi_{j+1},...,\xi_{n})\right\}=\frac{P_{\rm
c}}{n-1},
 \ j=1,...,n-1,
$$
$$
{\bf P}\{\xi'=\xi, \ \eta'=\eta\}=1-P_{\rm c}.
$$

In the {\em mutation operator}~${\rm Mut}$,
each bit of an offspring genotype may be flipped with a constant
mutation probability~$P_{\rm m}$, which is usually chosen
relatively small. When the whole population~$X^{t+1}$ of~$N$
offspring is constructed, the GA proceeds to the next
iteration~$t+1$. An initial population~$X^{0}$ is generated
randomly with independent choice of all bits in genotypes.

A plenty of variants of GA have been developed since publication
of the simple GA in~\cite{Holl75}, sharing the basic ideas, but
using different population management strategies, selection,
crossover and mutation operators~\cite{RR02}. The practice shows
that the best results are obtained when the GAs are designed in
view of the specific features of the optimization problem to be
solved. A number of such problem-specific GAs make use of
crossover operators that find exact or at least approximate
solution to the optimal recombination problem.

\subsection{Formulation of Optimal Recombination Problem}

In this paper, the ORPs are considered assuming binary
representation of solutions in genotypes being identical to the
solutions encoding of the NP~optimization problem. Besides that,
it will be assumed that~$X^0$ consists of feasible solutions and
operators ${\rm Cross}$ and ${\rm Mut}$ maintain feasibility of
solutions, \ie ${\rm Cross} : {\rm Sol}^{2} \to {\rm Sol}^{2}, \
{\rm Mut} : {\rm Sol} \to {\rm Sol}$. Therefore the term
"genotype" will mean an element of the set of feasible
solutions~${\rm Sol}$.


Note that there may be a number of NP~optimization problems,
essentially corresponding to the same problem in practice. Such
formulations are usually easy to transform to each other but the
solution representations may be quite different in the degree of
degeneracy, the number of local optima for some standard
neighborhood definitions, the length of encoding strings and other
parameters important for heuristic algorithms. Since the method of
solutions representation is crucial for recombination operators,
in what follows we will always explicitly indicate what solutions
encoding is used in formulation of an NP~optimization problem.

In general, an instance of an NP~optimization problem may have no
feasible solutions. However, w.r.t. the optimal recombination
problem such cases are not meaningful, \since there exist no
feasible parent solutions. Therefore, in the context of optimal
recombination below we will always assume that~${\rm Sol} \ne
\emptyset$.

The following definition of optimal recombination problem is
motivated by the principles of {\em (strictly) gene transmitting}
recombination formulated by N.~Radcliffe~\cite{R94}.

\begin{definition}\label{def:ORP}
Given an NP~optimization problem~$\Pi=(\mbox{\rm Inst},{\rm Sol},
f)$, the optimal recombination problem for~$\Pi$ is the
NP~optimization problem~$\overline{\Pi} =(\overline{\rm Inst},
\overline{\rm Sol},\overline{f})$, where for every instance
${\overline{I}=(I,{\bf p}^1,{\bf p}^2) \in \overline{\rm Inst}}$
holds $I \in {\rm Inst}$, ${{\bf p}^1=(p_1^1,\dots,p_{n(I)}^1)\in
{\rm Sol}(I)}, \ {{\bf p}^2=(p_1^2,\dots,p_{n(I)}^2) \in {\rm
Sol}(I)}$, and it is assumed that
\begin{equation}\label{eqn:condition_a}
\overline{\rm Sol}(\overline{I}) =
 \{\x \in {\rm
Sol}(I) | \ x_j =p^1_j \ \mbox{or} \ x_j =p^2_j, \
j=1,\dots,{n(I)}\}.
\end{equation}
The optimization criterion in~$\overline{I}$ is the same as
in~$I$, \ie~$\overline{f}_{\overline{I}} \equiv f_I$.
\end{definition}

The feasible solutions ${\bf p}^1,{\bf p}^2$ to problem~$I$ are
called the {\em parent} solutions for the
problem~$\overline{I}=(I,{\bf p}^1,{\bf p}^2)$. In what follows,
we denote the set of coordinates, where the parent solutions have
different values, by $D(\p^1,\p^2)=\{j: p^1_j \ne p^2_j\}.$ These
are the variables subject to optimization in the ORP. All other
variables are "fixed" in the ORP being equal to the values of the
corresponding coordinates in the parent solutions.


Other formulations of recombination subproblem, that may be found
in literature, are the examples of {\it allelic dynastically
optimal recombination}~\cite{C03}. In particular,
in~\cite{BDE,CS03,ErOR,LPP01} promising experimental results were
demonstrated by GAs where the recombination subproblem is defined
by "fixing" only those genes, where both parent genotypes contain
zeros.

\section{Efficiently Solvable Optimal Recombination Problems}\label{sec:PolySolve}

As the first examples of efficiently solvable ORPs we will
consider the following three well-known problems. Given a graph
$G=(V,E)$ with vertex weights $w(v), \ v \in V$,

\begin{itemize}
\item the Maximum Weight Independent Set Problem asks for a subset
${S\subseteq V}$, such that each edge~${e \in E}$ has at least one
endpoint outside~$S$ (i.e. $S$ is an independent set) and the
weight $\sum_{v \in S} w(v)$ of~$S$ is maximized;

\item the Maximum Weight Clique Problem asks for a maximum weight
subset ${Q\subseteq V}$, such that any two vertices~$u,v$ in~$Q$
are adjacent (\ie $Q$ is a clique);

\item the Minimum Weight Vertex Cover Problem asks for a minimum
weight subset ${C\subseteq V}$, such that any edge ${e \in E}$ is
incident at least to one of the vertices in~$C$ (\ie $C$ is a
vertex cover).
\end{itemize}

Suppose, the vertices of graph $G$ are ordered. We will consider
these three problems using the standard binary representation of
solutions by the indicator vectors, assuming $n=|V|$ and $x_j=1$
iff vertex~$v_j$ belongs to the subset represented by~$\x$. The
following result is due to E.~Balas and W.~Niehaus.

\begin{theorem}\label{AOT} {\rm \cite{BN95}} The ORP for
the Maximum Weight Clique Problem is solvable in
time~${O(|D(\p^1,\p^2)|^3+n)}$.
\end{theorem}

{\bf Proof.} Consider the Maximum Weight Clique Problem on a given
graph~$G$ with two parent cliques~$Q_1$ and $Q_2$, represented by
binary vectors~$\p^1$ and $\p^2$. An offspring solution~$Q$ should
contain the whole set of vertices $Q_1\cap Q_2$, besides that~$Q$
should not contain the elements from the set $V\setminus (Q_1\cup
Q_2)$, while the vertices with indices from the set~$D({\bf
p}^1,{\bf p}^2)$ should be chosen optimally. The latter task can
be formulated as a Maximum Weight Clique Problem in
subgraph~$H=(V',E')$, which is induced by the subset of vertices
with indices from~$D({\bf p}^1,{\bf p}^2)$. To find a clique of
maximum weight in~$H$, it is sufficient to find a minimum weight
vertex cover~$C'$ in the complement graph~$\bar{H}$ and take
$V'\backslash C'$. Note that~$\bar{H}$ is a bipartite graph, so
let~$V'_1,V'_2$ be the subsets of vertices in this bipartition.

The Minimum Weight Vertex Cover~$C'$ for~$\bar{H}$ can be found by
solving the $s$-$t$-Minimum Cut Problem on a supplementary
network~${\cal N}$, based on~$\bar{H}$, as described e.g.
in~\cite{Hoch97}: in this network, an additional vertex~$s$ is
connected by outgoing arcs with the vertices of set~$V'_1$, and
the other additional vertex~$t$ is connected by incoming arcs to
the subset~$V'_2$. The capacities of the new arcs are equal to the
weights of the adjacent vertices in~$\bar{H}$. Each edge
of~$\bar{H}$ is viewed as an arc, directed from its endpoint $u
\in V_1'$ to the endpoint $v \in V_2'$. The arc capacity is set to
$\max\{w(u),w(v)\}$. This $s$-$t$-Minimum Cut Problem can be
solved in~${O(|D(\p^1,\p^2)|^3)}$ time using the maximum-flow
algorithm due to A.V.~Karzanov -- see e.g.~\cite{PS82}.
We will assume that the $s$-$t$-minimum cut contains only the arcs
outgoing from~$s$ or incoming into~$t$, because if some arc
$(u,v),$ $u\in V'_1, v\in V'_2$ enters the $s$-$t$-minimum cut,
one can substitute it by $(s,u)$ or $(v,t)$, and this will not
increase the weight of the cut.
Finally, it is easy to verify that~$(V'_1 \cup V'_2) \backslash
C'$ joined with $Q_1\cap Q_2$ defines the required ORP solution.
Since the parent solutions are given by the $n$-dimensional
indicator vectors~$\p^1$ and $\p^2$, we get the overall time
complexity~${O(|D(\p^1,\p^2)|^3+n)}$.
 $\Box$\\

%
%
%
%
%
%
%
%

Note that if all vertex weights are equal, then the time
complexity of Karzanov's algorithm for the networks of simple
structure (as the one constructed in the proof of
Theorem~\ref{AOT}) reduces to ${O(|D(\p^1,\p^2)|^{2.5})}$ --
see~\cite{PS82}.

The Maximum Weight Independent Set and the Minimum Weight Vertex
Cover Problems are closely related to the Maximum Weight Clique
Problem (see e.g.~\cite{GJ}). It is sufficient to consider the
complement graph and to change the optimization criterion
accordingly. Then there is a bijection between the set of feasible
solutions of each of these problems and the set of feasible
solutions of the corresponding Maximum Weight Clique Problem. In
the case of Maximum Weight Independent Set, the bijection is an
identity mapping, while in the case of the Minimum Weight Vertex
Cover, the bijection alters each bit in~$\x$. In the first case
the mapped feasible solutions retain thir objective function
values, while in the second case the original objective function
values are subtracted from the weight of all vertices. In view of
these relationships Theorem~\ref{AOT} implies that the ORPs for
the Maximum Weight Independent Set and the Minimum Weight Vertex
Cover Problems are solvable in time~${O(|D(\p^1,\p^2)|^3+n)}$ as
well. Indeed, it suffices to consider the corresponding instance
of the ORP for the Maximum Clique Problem, solve this ORP in
${O(|D(\p^1,\p^2)|^3+n)}$ time and map the obtained solution back
into the set of feasible solutions of the original problem.

The above arguments illustrate that when one NP-optimization
problem transforms efficiently to another one, the corresponding
ORPs may reduce efficiently as well. The following subsection is
devoted to analysis of the situations where such arguments apply.

\subsection{Reductions of Optimal Recombination Problems}

The usual approach to spreading a class of polynomially solvable
(or intractable) problems consists in building chains of efficient
problem reductions. In order to apply this approach to optimal
recombination problems we shall first formulate a relatively
general reducibility condition for NP~optimization problems.

\begin{proposition} \label{priznak}
Let ${\Pi_1={({\rm Inst}_1,{\rm Sol}_1,f_I)}}$ and $\Pi_2={({\rm
Inst}_2,{\rm Sol}_2,g_{I'})}$ be NP~optimization problems with
maximization (minimization) criteria and there exists a
mapping~${\alpha: {\rm Inst}_1 \to {\rm Inst}_2}$ and an injective
mapping~${\beta \ :\ {\rm Sol}_1(I) \to {\rm Sol}_2(\alpha(I))}$,
such that given~$I \in {\rm Inst}_1$,
\begin{enumerate}
\item
for any $\x,\x' \in {\rm Sol}_1(I)$, satisfying the condition
\begin{equation} \label{eqn:f_I}
f_{I}(\x)
> f_{I}(\x'),
\end{equation}
the following inequality holds
\begin{equation} \label{eqn:g_alphaI}
g_{\alpha(I)}(\beta(\x)) > g_{\alpha(I)}(\beta(\x'))
\end{equation}
(if $\Pi_1$ is a minimization problem, the inequality sign
in~(\ref{eqn:f_I}) changes into~"$<$"; if $\Pi_2$ is a
minimization problem, the inequality sign in~(\ref{eqn:g_alphaI})
changes into~"$<$");
\item
if $\y \in \beta({\rm Sol}_1(I))$, $\y' \in {\rm
Sol}_2(\alpha(I))$, and
\begin{equation}\label{eqn:good_enough}
g_{\alpha(I)}(\y') \ge g_{\alpha(I)}(\y),
\end{equation}
then $\y' \in \beta({\rm Sol}_1(I))$
(if $\Pi_2$ is a minimization problem, the inequality sign
in~(\ref{eqn:good_enough}) changes into~"$\le$"
).

\end{enumerate}
Then $\Pi_1$ transforms to $\Pi_2$, so that any instance $I \in
{\rm Inst}_1$ can be solved in
time~$O(T_{\alpha}(I)+T_{\beta^{-1}}(I)+T(I))$, where
$T_{\alpha}(I)$ is the computation time of~$\alpha(I)$;
$T_{\beta^{-1}}(I)$ is an upper bound on the computation time of
$\beta^{-1}(\y),$ $\y \in \beta({\rm Sol}_1(I))$; $T(I)$ is the
time complexity of solving the problem~$\alpha(I)$.
\end{proposition}

{\bf Proof.} Suppose~$I \in {\rm Inst}_1$ and consider an optimal
solution~$\y^*$ to problem~$\alpha(I)$. According to condition~2,
if ${\rm Sol}_1(I) \ne \emptyset$, then $\y^*\in \beta({\rm
Sol}_1(I))$. By proof from the contrary, in view of condition~1,
we conclude that if~${\rm Sol}_1(I) \ne \emptyset$, then
$\beta^{-1}(\y^*)$ is an optimal solution to~$I$. $\Box$\\

Note that condition~2 in Proposition~\ref{priznak} implies that
the set of feasible solutions of problem~$\Pi_1$ is mapped into a
set of "sufficiently good" feasible solutions to~$\Pi_2$ (in terms
of objective function). This property is observed in many
transformations involving penalization of "undesired" solutions
to~$\Pi_2$ (see e.g.~\cite{BGD,KP83}).

%

If the computation times $T_{\alpha}(I)$ and $T_{\beta^{-1}}(I)$
are polynomially bounded w.r.t.~$|I|$, then
Proposition~\ref{priznak} provides a sufficient condition of
polynomial reducibility of one NP~optimization to another.

The following proposition is aimed at obtaining efficient
reductions of one ORP to another, when there exist efficient
transformations between the corresponding NP~optimization
problems.

\begin{proposition}\label{reduction0}
Let ${\Pi_1=({\rm Inst}_1,{\rm Sol}_1,f_{I})}$ and ${\Pi_2=({\rm
Inst}_2, {\rm Sol}_2,g_{I'})}$ be both NP~optimization problems,
where ${\rm Sol}_1(I)\subseteq \{0,1\}^{n_1(I)}$, ${\rm
Sol}_2(I')\subseteq \{0,1\}^{n_2(I')}$ and there exist the
mappings $\alpha$ and $\beta$ for which the condition of
Proposition~\ref{priznak} holds and besides that:

(i) For any $j=1,\dots,n_1(I)$ there exists such $k(j)$ that
$\beta^{-1}(\y)_{j}$ is a function of $y_{k(j)}$, when ${\bf
y}=(y_1,\dots,y_{n_2}) \in \beta({\rm Sol}_1(I))$.

(ii) For any $k=1,\dots,n_2(\alpha(I))$ there exists such~$j(k)$
that $\beta(\x)_{k}$ is a function of $x_{j(k)}$, when ${\bf
x}=(x_1,\dots,x_{n_1}) \in {\rm Sol}_1(I)$.

Then $\overline{\Pi}_1$ reduces to $\overline{\Pi}_2$, and any
instance $\overline{I}=(I,\p^1,\p^2)$ from ORP~$\overline{\Pi}_1$
is solvable in time
$O(T_{\alpha}(I)+T_{\beta}(I)+T_{\beta^{-1}}(I)+\overline{T}(I,\p^1,\p^2))$,
where $\overline{T}(I,\p^1,\p^2)$ is the time complexity of
solving ORP~$(\alpha(I),\beta(\p^1),\beta(\p^2))$, and
$T_{\beta}(I)$ is an upper bound on computation time
of~$\beta(\x), \ \x \in {\rm Sol}_1(I)$.
\end{proposition}

{\bf Proof.} Without loss of generality we shall assume
that~$\Pi_1$ and $\Pi_1$ are maximization problems. Suppose, an
instance~$I$ of problem~$\Pi_1$ and two parent solutions
$\p^1,\p^2 \in {\rm Sol}_1(I)$ are given. These solutions
correspond to feasible solutions ${\bf q}^1=\beta({\bf p}^1)$,
${\bf q}^2=\beta({\bf p}^2)$ to problem~$\alpha(I)$.

Now let us consider the ORP for instance $\alpha(I)$ of $\Pi_2$
with parent solutions~${\bf q}^1, {\bf q}^2$. Optimal solution to
this ORP ${\bf y'}\in {\rm Sol}_2(\alpha(I))$ can be transformed
in time~$T_{\beta^{-1}}$ into a feasible
solution~$\z=\beta^{-1}(\y') \in {\rm Sol}_1(I)$.

Note that for all $j \not \in D(\p^1,\p^2)$ hold
$z_j=p^1_j=p^2_j$. Indeed, by condition~(i), for any
$j=1,\dots,n_1(I)$ there exists such $k(j)$ that

(I)~either $\beta^{-1}({\bf y})_{j}=y_{k(j)}$ for all ${\bf y} \in
\beta({\rm Sol}_1(I))$, or

(II)~$\beta^{-1}({\bf y})_{j}=1-y_{k(j)}$ for all ${\bf y} \in
\beta({\rm Sol}_1(I))$, or

(III)~$\beta^{-1}({\bf y})_{j}$ is constant on~$\beta({\rm
Sol}_1(I))$.

In the case~(I) for all $j \not \in D(\p^1,\p^2)$ we have
$z_j=y'_{k(j)}$. Now $y'_{k(j)} = q^1_{k(j)}$ by the definition of
the ORP, since $q^1_{k(j)}=p^1_j=p^2_j=q^2_{k(j)}$. So,
$z_j=q^1_{k(j)}=p^1_j=p^2_j.$ The case~(II) is treated
analogously. Finally, the case~(III) is trivial since
$\z,\p^1,\p^2 \in \beta^{-1}(\beta({\rm Sol}_1(I)))$. So, $\z$ is
a feasible solution to the ORP for~$\Pi_1$.

To prove the optimality of~$\z$ for instance~$\overline{I}$ from
the ORP~$\overline{\Pi}_1$ we will assume by contradiction that
there exists a feasible solution~${\bf z}'=(z'_1,\dots,z'_{n_1})
\in {\rm Sol}_1(I)$ such that ${\bf z}'_j=p^1_j=p^2_j$ for all $j
\not\in D(\p^1,\p^2)$, and $f_I({\bf z}')>f_I({\bf z}')$. Then
$g_{\alpha(I)}(\beta({\bf z}')) >
g_{\alpha(I)}(\beta(\z))=g_{\alpha(I)}(\y')$. But $\beta({\bf
z}')$ coincides with~$\q^1$ and $\q^2$ in all coordinates
$k\not\in D(\q^1,\q^2)$ according to condition~(ii) (it is
sufficient to consider three cases similar to~(I) -- (III) in
order to verify this). Thus~$\y'$ is not an optimal solution to
the ORP for~$\alpha(I)$, which is a contradiction. $\Box$\\

The special case of this proposition where $n_1(I)\equiv n_2(I')$
and $k(j)\equiv j, \ j(k)\equiv k$ appears to be the most
applicable, as it is demonstrated in what follows.


Let us use Proposition~\ref{reduction0} to obtain an efficient
optimal recombination algorithm for the {\em Maximum Weight Set
Packing Problem:}
\begin{equation}\label{SPP}
\max\left\{f_{\rm pack}(\x)={\bf cx} : {\bf Ax} \le {\bf e}, \x
\in \{0,1\}^n\right\},
\end{equation}
where ${\bf A}$ is a given $(m\times n)$-matrix of zeros and ones.
Here and below ${\bf e}$ is an $m$-dimensional column vector of
ones. The transformation~$\alpha$ from the Set Packing to the
Maximum Weight Independent Set Problem (with the standard binary
solutions encoding) consists in building a graph on a set of
vertices $v_1,\dots,v_n$ with weights $c_1,\dots,c_n$. Each pair
of vertices $v_j, v_k$ is connected by an edge iff $j$ and $k$
both belong at least to one of the subsets $N_i = \{j: a_{ij} \ne
0\}.$ In this case $\beta$ is an identical mapping. Application of
Proposition~\ref{reduction0} leads to

\begin{corollary}\label{setpack} {\rm \cite{ErECJ08}} The ORP for the Maximum
Weight Set Packing Problem {\em(\ref{SPP})} is solvable in
time~$O(|D(\p^1,\p^2)|^3+n^2 m)$.
\end{corollary}

Now we can prove the polynomial solvability of the next two
problems in Boolean linear programming formulations.

The first problem is the {\em Minimum Weight Set Partition
Problem:}
\begin{equation}\label{SPartP}
\min\left\{f_{\rm part}(\x)={\bf cx} : {\bf Ax} = {\bf e}, \x \in
\{0,1\}^n\right\},
\end{equation}
where ${\bf A}$ is a given $(m\times n)$-matrix of zeros and ones.

The second problem is the {\em Simple Plant Location Problem.}
Suppose there are~$n$ sites of potential facility location for
production of some uniform product. The cost of opening a facility
at location~$i$ is~$C_i \geq 0$. Each open facility can provide an
unlimited amount of commodity.

Suppose there are~$m$ customers that require service and the cost
of serving a client~$j$ by facility~$i$ is~$c_{ij}\geq 0$. The
goal is to determine a set of sites where the facilities should be
opened so as to minimize the total opening and service cost. This
problem can be formulated as a nonlinear Boolean Programming
Problem:
\begin{equation}\label{SPPLgoal0}
\mbox{min} \ \ f_{\rm splp}({\bf x})=\sum\limits_{i=1}^n C_i x_i +
  \sum\limits_{j=1}^m \min \limits_{i: x_i=1} c_{ij},
\end{equation}
s.~t.
\begin{equation}\label{SPPL_nonempty}
\sum\limits_{i=1}^n x_i \ge 1.
\end{equation}
Here the vector of variables ${\bf x}=(x_1,\dots,x_n) \in
\{0,1\}^n$ is an indicator vector for the set of opened
facilities. Note that given a vector of open facilities, a least
cost assignment of clients to these facilities is easy to find. An
optimal solution to the Simple Plant Location Problem in the above
formulation is denoted by~${\bf x}^*$.

The Simple Plant Location Problem is strongly NP-hard even if the
matrix~$(c_{ij})$ satisfies the triangle inequality~\cite{KV05}.
Interconnections of this problem to other well-known optimization
problems may be found in~\cite{BGD,KP83} and the references
provided there.

Alternatively, the Simple Plant Location Problem may be formulated
as a Boolean Linear Programming Problem:

\begin{equation}\label{SPPLgoal}
\mbox{min} \ \  f_{\rm splp}({\bf Y},{\bf u}) = \sum_{k=1}^K
\sum_{\ell=1}^{L} c_{k\ell}y_{k\ell} +
    \sum_{k=1}^K C_{k}u_{k},
\end{equation}
\begin{equation}\label{SPPLserve}
\sum_{k=1}^{K} y_{k\ell} = 1, \quad \ell=1,\dots,L,
\end{equation}
\begin{equation}\label{SPPLopen}
u_k \ge y_{k\ell}, \quad k=1,\dots,K, \ \ell=1,\dots,L,
\end{equation}
\begin{equation}\label{SPPLbool}
y_{k\ell} \in \{0,1\}, \ u_{k}\in \{0,1\}, \quad k=1,\dots,K, \
\ell=1,\dots,L.
\end{equation}
Here and below, we denote the ($K\times L$)-matrix of Boolean
variables~$y_{k\ell}$ by ${\bf Y}$, and the $K$-dimensional vector
of Boolean variables~$u_k$ is denoted by~${\bf u}$. This
formulation of the Simple Plant Location Problem is equivalent
to~(\ref{SPPLgoal0}) -- (\ref{SPPL_nonempty}). However, according
to Definition~\ref{def:NPO}, the NP~optimization
problem~(\ref{SPPLgoal0})--(\ref{SPPL_nonempty}) is different from
problem~(\ref{SPPLgoal})--(\ref{SPPLbool}), since in the first
case the feasible solutions are encoded by vectors~$\x \in
\{0,1\}^n$ while in the second case the feasible solutions are
encoded by pairs~$({\bf Y},{\bf u})$.

On one hand, in Section~\ref{sec:NPHard} it will be shown that the
ORP for the Simple Plant Location
Problem~(\ref{SPPLgoal0})--(\ref{SPPL_nonempty}) is NP-hard. On
the other hand, the following corollary shows that the ORP for
Simple Plant Location Problem~(\ref{SPPLgoal})--(\ref{SPPLbool})
is efficiently solvable, as well as the ORP for the Set Partition
Problem~(\ref{SPartP}).

\begin{corollary}\label{setpart_splp} {\rm \cite{ErECJ08}}

(i) The ORP for the Minimum Weight Set Partition
Problem~(\ref{SPartP}) is solvable in time~$O(|D(\p^1,\p^2)|^3+n^2
m)$.

(ii) The ORP for the Simple Plant Location Problem in Boolean
Linear Programming formulation~(\ref{SPPLgoal})--(\ref{SPPLbool})
is solvable in polynomial time.
\end{corollary}

{\bf Proof.} For both cases we will use the well-known
transformations of the corresponding NP~optimization problems to
the Minimum Weight Set Packing Problem (see e.g. the
transformations~T2 and T5 in~\cite{KP83}).

(i) Let us denote the Minimum Weight Set Partition Problem
by~$\Pi_1$ and let the Set Packing Problem be~$\Pi_2$. Since ${\rm
Sol}_1(I)\ne \emptyset$, the problem~$I$ is equivalent to
$$
\mbox{min} \ \ \sum_{j=1}^n c_j x_j + \lambda\sum_{i=1}^m w_i,
$$
subject to
$$
\sum_{j=1}^n a_{ij} x_j + w_i=1, \ i=1,\dots,m,
$$
$$
x_j \in \{0,1\}, \ j=1,\dots,n; \ \ w_i\ge 0, \ i=1,\dots,m,
$$
where $\lambda>2\sum_{j=1}^n |c_j|$ is a penalty factor which
assures that all "artificial" slack variables~$w_i$ become zeros
in the optimal solution. By substitution of~$w_i$ into the
objective function, the latter model transforms into
$$
\min\left\{\lambda m+\sum_{j=1}^n\left(c_j-\lambda\sum_{i=1}^m
a_{ij}\right)x_j : \A\x \le \e, \ \x \in \{0,1\}^n\right\},
$$
which is equivalent to the following instance~$\alpha(I)$ of the
Set Packing Problem~$\Pi_2$:
$$
\max\left\{g(\x)=\sum_{j=1}^n\left(\lambda\sum_{i=1}^m a_{ij} -c_j
\right)x_j : \A\x \le \e, \ \x \in \{0,1\}^n\right\}.
$$

Assume that~$\beta$ is an identical mapping. Then each feasible
solution~$\x$ of the Set Partition Problem is a feasible solution
to problem~$\Pi_2$ with the objective function value
$g(\x)=\lambda m - f_{\rm part}(\x) > \lambda (m-1/2).$ At the
same time, if a vector~$\x'$ is feasible for problem~$\Pi_2$ but
infeasible for~$\Pi_1$, it will have the objective function value
$g(\x')=\lambda (m-k)-f_{\rm part}(\x'),$ where $k$ is the number
of constraints of the form $\sum_{j=1}^n a_{ij} x_j=1,$ which are
violated by~$\x'$. So, $\beta$ is a bijection from ${\rm
Sol}_1(I)$ to a set of feasible solutions with sufficiently high
values of the objective function:
$$
\{ {\bf x} \in {\rm Sol}_2(\alpha(I)) \ | \ g({\bf x}) > \lambda
(m-1/2)\}.
$$

The complexity of ORP for~$\Pi_2$ is bounded by
Corollary~\ref{setpack}. Thus, application of
Proposition~\ref{reduction0} completes the
proof of part~(i).\\

(ii) Let~$\Pi'_1$ be the Simple Plant Location Problem.
Analogously to the case~(i) we will convert
equations~(\ref{SPPLserve}) into inequalities. To this end, we
rewrite~(\ref{SPPLserve}) as $\sum_{k=1}^K y_{k\ell}+w_{\ell}= 1,\
\ell=1,\dots,L,$ with nonnegative slack variables~$w_{\ell}$ and
ensure all of them turn into zero in the optimal solution, by
means of a penalty term $\lambda \sum_{\ell=1}^L w_{\ell}$ added
to the objective function. Here
$$
\lambda>\sum_{k=1}^K C_k + \sum_{\ell=1}^L \max_{k=1,\dots,K}
c_{k\ell}.
$$
Eliminating variables~$w_{\ell}$ we substitute~(\ref{SPPLserve})
by $\sum_{k=1}^K y_{k\ell} \le 1,\ \ell=1,\dots,L,$ and change the
penalty term into $\lambda L - \lambda \sum_{\ell=1}^L
\sum_{k=1}^K y_{k\ell}$. Multiplying the criterion by~$-1$ and
introducing a new set of variables $\overline{u}_k = 1-u_k, \
k=1,\dots,K$, we obtain the following~NP~maximization
problem~$\Pi'_2$:
\begin{equation}\label{P2goal}
\mbox{max} \ \    g'({\bf Y},\overline{\bf u}) = \sum_{k=1}^K
\sum_{\ell=1}^{L} (\lambda-c_{k\ell})y_{k\ell}
   +
    \sum_{k=1}^K C_{k}\overline{u}_{k} - \lambda L - \sum_{k=1}^K C_k,
\end{equation}
subject to
\begin{equation}\label{P2serve}
\sum_{k=1}^{K} y_{k\ell} \le 1, \quad \ell=1,\dots,L,
\end{equation}
\begin{equation}\label{P2open}
\overline{u}_k + y_{k\ell} \le 1, \quad k=1,\dots,K, \
\ell=1,\dots,L,
\end{equation}
\begin{equation}\label{P2bool}
y_{k\ell} \in \{0,1\}, \ \overline{u}_{k}\in \{0,1\}, \quad
k=1,\dots,K, \ \ell=1,\dots,L,
\end{equation}
where $\overline{\bf u}=(\overline{u}_{1},\dots,
\overline{u}_{K})$. Obviously, $\Pi'_2$ is a special case of the
Set Packing Problem, up to an additive constant~$-\lambda L -
\sum_{k=1}^K C_k$ in the objective function. Thus, we have defined
the mapping~$\alpha(I)$.

Assume that~$\beta$ maps identically all variables $y_{k\ell}$ and
transforms the variables~$u_k$ into $\overline{u}_k = 1-u_k, \
k=1,\dots,K$. Then each feasible solution $({\bf Y},{\bf u})$ of
the Simple Plant Location Problem is mapped into a feasible
solution to problem~$\Pi'_2$ with an objective function value
$g'({\bf Y},\overline{\bf u})=- f_{\rm splp}({\bf Y},{\bf
u})>-\lambda$. If a pair~$({\bf Y},\overline{\bf u})$ is feasible
for problem~$\Pi'_2$ but $({\bf Y},{\bf u})$ is infeasible
in~$\Pi'_1$, then $g'({\bf Y},\overline{\bf u}) \le - f_{\rm splp}
({\bf Y},{\bf u}) - \lambda,$ because at least one of the
equalities~(\ref{SPPLserve}) is violated by~$({\bf Y},{\bf u})$.

The ORP for the problem~$\Pi'_2$ can be solved in polynomial time
by Corollary~\ref{setpack}, thus Proposition~\ref{reduction0}
gives
the required optimal recombination algorithm for~$\Pi'_1$. $\Box$\\



\begin{figure}
\begin{center}
\includegraphics[height=6.33cm,width=9.85cm]{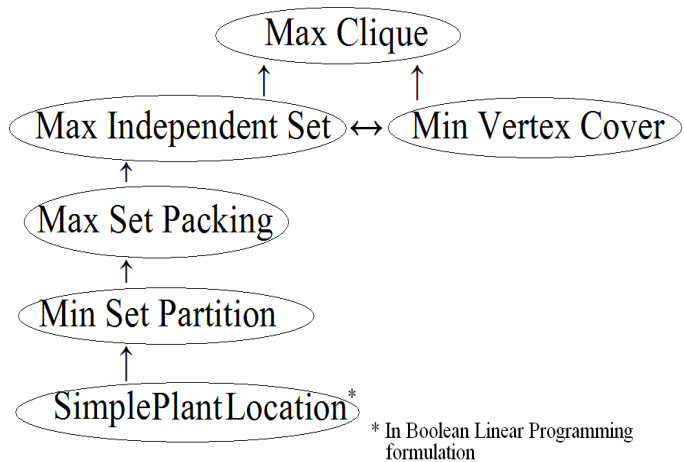}
\end{center}
\vspace{-1em} \caption{Polynomial-time reductions of optimal
recombination problems (all displayed problems are in weighted
versions).}\label{fig:reductions}
\end{figure}

The ORP reductions described above are illustrated in
Fig.~\ref{fig:reductions}.

\subsection{Boolean Linear Programming Problems and Hypergraphs}

The starting point of all reductions considered above was
Theorem~\ref{AOT} which may be viewed as an efficient reduction of
the ORP for the Maximum Weight Clique Problem to the Maximum
Weight Independent Set Problem in a bipartite graph. In order to
generalize this approach now we will move from bipartite graphs to
2-colorable hypergraphs.

A hypergraph $H=(V,E)$ is given by a finite nonempty set of
vertices~$V$ and a set of edges~$E$, where each edge~$e\in E$ is a
subset of~$V$. A subset $S\subseteq V$ is called {\em independent}
if none of the edges $e \in E$ is a subset of~$S$. The Maximum
Weight Independent Set Problem on hypergraph $H=(V,E)$ with
rational vertex weights $w(v), \ v \in V$ asks for an independent
set $S$ with maximum weight $w(S)=\sum_{v \in S} w(v)$.

A generalization of the bipartite graph is the {\em 2-colorable
hypergraph}: there exists a partition of the vertex set~$V$ into
two disjoint independent subsets $C_1$ and $C_2$. The partition
$V=C_1\cup C_2$, $C_1 \cap C_2=\emptyset$ is called a 2-coloring
of $H$ and $C_1, C_2$ are the color classes.

Let us denote by~$N_i$ the set of indices of non-zero elements in
constraint~$i$ of the Boolean Linear Programming
Problem~(\ref{goal})-(\ref{bools}). In the sequel we will assume
that at least one of the subsets~$N_i$ contains two or more
elements (otherwise the problem is solved trivially).

\begin{theorem}\label{hyper2} {\rm \cite{ErECJ08}}
The ORP for Boolean Linear Programming
Problem~(\ref{goal})-(\ref{bools}) reduces to the Maximum Weight
Independent Set Problem on a 2-colorable hypergraph with a
2-coloring given in the input. Each edge in the 2-colorable
hypergraph contains at most~$N_{\max}$ vertices, where
$N_{\max}=\max_{i=1,\dots,m} |N_i|$, and the time complexity of
this reduction is~$O(m(2^{N_{\max}}+n))$.
\end{theorem}

{\bf Proof.} Given an instance of the Boolean Linear Programming
Problem with parent solutions~$\p^1$ and~$\p^2$, let us denote
$|D(\p^1,\p^2)|$ by~$d$ and construct a hypergraph~$H$ on~$2d$
vertices, assigning each variable $x_j,\ j\in D(\p^1,\p^2),$ a
couple of vertices $v_j$ and $v_{n+j}$. In order to model each of
the linear constraints for $i=1,\dots,m$ we will look through all
possible combinations of the Boolean variables from $D(\p^1,\p^2)$
involved in this constraint:
$$
\{\x \in \{0,1\}^n: x_j=0 \ \forall j \not\in N_i\cap
D(\p^1,\p^2)\}.
$$
Let $\x^{ik}, k=1,\dots,2^{|N_{i} \cap D(\p^1,\p^2)|}$ denote the
$k$-th vector in this set. For each combination~$k$ which violates
a constraint~$i$ from~(\ref{ineq}), i.e.
$$
\sum_{j\in N_i\cap D(\p^1,\p^2)} a_{ij} x^{ik}_j
 +
\sum_{j\in N_i \backslash D(\p^1,\p^2)} a_{ij} p^{1}_j > b_i,
$$
we add an edge
$$
e_{ik}=\{v_j: x^{ik}_j=1, \ j \in N_i\cap D(\p^1,\p^2)\} \cup
 \{v_{j+n}: x^{ik}_j=0, \ j \in N_i\cap D(\p^1,\p^2)\}
$$
into the hypergraph. (Note that the edge~$e_{ik}$ contains at
most~$|N_i|$ elements.) Besides that, we add~$d$ edges
$\{v_j,v_{n+j}\}, j\in D(p_1,p_2)$, to guarantee that both $v_j$
and~$v_{n+j}$ can not enter into an independent set together.

If~$\x$ is a feasible solution to the ORP
for~(\ref{goal})-(\ref{bools}), then the set of vertices
$$
S(\x)=\{v_j: x_j=1, j\in D(p_1,p_2)\} \cup \{v_{j+n}: x_j=0, j\in
D(p_1,p_2)\}
$$
is independent in~$H$. Given a set of vertices~$S$, we can
construct the corresponding vector~$\x(S)$, assigning $\x(S)_j=1$
if $v_j \in S,$ $j\in D(\p^1,\p^2)$ or if $p^1_j=p^2_j=1$.
Otherwise $\x(S)_j=0.$ Then for each independent set~$S$ of~$d$
vertices, $\x(S)$ is feasible in the Boolean Linear Programming
Problem.

The hypergraph vertices are given the following weights:
$$
w(v_j)=c_j+\lambda, \ w(v_{n+j})=\lambda, \ {j\in D(\p^1,\p^2),}
$$
where $\lambda>2\sum_{j\in D(p_1,p_2)} |c_j|$.

Now each maximum weight independent set~$S^*$ contains
either~$v_j$ or~$v_{n+j}$ for any ${j\in D(\p^1,\p^2)}$. Indeed,
there must exist a feasible solution to the ORP and it corresponds
to an independent set of weight at least~$\lambda d$. However, if
an independent set neither contains~$v_j$ nor~$v_{n+j}$ then its
weight is below $\lambda d - \lambda/2$.

So, optimal independent set~$S^*$ corresponds to a feasible
vector~$\x(S^*)$ with the goal function value
$$
{\bf cx}(S^*)=
 \sum_{j\in S^*, \ j\le n} c_{j}
 +
 \sum_{j\not\in D(\p^1,\p^2)} c_{j} p^{1}_j
 =
 w(S^*)-\lambda d
 +
 \sum_{j\not\in D(\p^1,\p^2)} c_{j} p^{1}_j.
$$
Under the mapping~$S(\x)$, which is inverse to~$\x(S)$, any
feasible vector~$\x$ yields an independent set of weight
$$
w(S(\x))={\bf cx}+\lambda d -  \sum_{j\not\in D(\p^1,\p^2)} c_{j}
p^{1}_j,
$$
therefore~$\x(S^*)$ is an optimal solution to the ORP. $\Box$\\

Note that if an edge $e \in H$ consists of a single vertex,
$e=\{v\}$, then the vertex~$v$ can not enter into the independent
sets. All of such vertices should be excluded from the
hypergraph~$H$ constructed in Theorem~\ref{hyper2}. Let us denote
the resulting hypergraph by~$H'$. If $N_{\max}\le 2$, then the
hypergraph~$H'$ is an ordinary graph with at most $2d$ vertices.
Thus, by Theorem~\ref{hyper2} the ORP reduces to the Maximum
Weight Independent Set Problem in a bipartite graph~$H'$, which is
solvable in~$O(d^3)$ operations. Using this fact,
Theorem~\ref{AOT} may be extended as follows:

\begin{corollary}\label{ip2} {\rm \cite{ErECJ08}} The ORP for Linear Boolean Programming
Problem with at most two variables per inequality is solvable in
time~$O(|D(\p^1,\p^2)|^3+mn)$, if the solutions are represented by
vectors $x \in \{0,1\}^n$.
\end{corollary}

The class of Linear Boolean Programming Problems with at most two
variables per inequality includes the Vertex Cover Problem and the
{\em Minimum 2-Satisfiability Problem} -- see e.g~\cite{Hoch97}.

\section{NP-hard Optimal Recombination Problems in
Boolean Linear Programming \label{sec:NPHard}}

It was shown above that the optimal recombination on the class of
Boolean Linear Programming Problems is related to the Maximum
Weight Independent Set Problem on hypergraphs with a given
2-coloring. The next lemma indicates that in general case the
latter problem is NP-hard.

\begin{lemma}\label{HyperIS_NP} {\rm \cite{ErECJ08}}
The problem of finding a maximum size independent set in a
hypergraph with all edges of size~3 is strongly NP-hard even if a
2-coloring is given.
\end{lemma}
{\bf Proof.} Let us construct a reduction from the strongly
NP-hard Maximum Size Independent Set Problem on ordinary graphs to
the problem under consideration. Given a graph~$G=(V,E)$ with the
set of vertices $V=\{v_1,\dots,v_{n}\}$, consider a
hypergraph~$H=(V',E')$ on the set of vertices
$V'=\{v_1,\dots,v_{2n}\}$, where for each edge $e=\{v_i,v_j\}\in
E$ there are $n$ edges of the form $\{v_i,v_j,v_{n+k}\}, \
k=1,\dots,n$ in~$E'$. A 2-coloring for this hypergraph can be
composed of color classes $C_1=V$ and
$C_2=\{v_{n+1},\dots,v_{2n}\}$. Any maximum size independent set
in this hypergraph consists of the set of vertices
$\{v_{n+1},\dots,v_{2n}\}$ joined with a maximum size independent
set~$S^*$ in~$G$. Therefore, any maximum size independent set
in~$H$ immediately induces a maximum size
independent set for~$G$. $\Box$\\

The Maximum Size Independent Set Problem in a hypergraph $H=(V,E)$
may be formulated as a Boolean Linear Programming Problem
\begin{equation}\label{hyperBool}
\max\left\{\mbox{$\sum_{j=1}^n x_j$} : {\bf Ax} \le {\bf b}, \x
\in \{0,1\}^n\right\}
\end{equation}
with $m=|E|, n=|V|,$ $b_i=|e_i|-1, \ i=1,\dots,m$ and $a_{ij}=1$
iff $v_j \in e_i$, otherwise $a_{ij}=0$. In the special case
where~$H$ is 2-colorable, we can take $\p^1$ and $\p^2$ as the
indicator vectors for the color classes~$C_1$ and~$C_2$ of any
2-coloring. Then $D(\p^1,\p^2)=\{1,\dots,n\}$ and the ORP for the
Boolean Linear Programming Problem~(\ref{hyperBool}) becomes
equivalent to solving the maximum size independent set in a
hypergraph~$H$ with a given 2-coloring. In view of
Lemma~\ref{HyperIS_NP}, this leads to the following

\begin{theorem} {\rm \cite{ErECJ08}}
The optimal recombination problem for Boolean Linear Programming
Problem is strongly NP-hard even in the case where $|N_i|=3$ for
all $i=1,\dots,m$; $c_j=1$ for all $j=1,\dots,n$ and matrix~${\bf
A}$ is Boolean.
\end{theorem}

In the rest of this section we will discuss NP-hardness of the
ORPs for some well-known Boolean Linear Programming Problems.

\subsection{One-Dimensional Knapsack and Bin Packing}

In Boolean linear programming formulation the One-Dimensional
Knapsack Problem has the following formulation

\begin{equation}\label{knap}
\max\left\{{\bf cx} : {\bf a} \x \le A, \x \in \{0,1\}^n\right\},
\end{equation}
where ${\bf c}=(c_1,\dots,c_n)$, ${\bf a}=(a_1,\dots,a_n)$, $a_j
\ge 0, c_j \ge 0, j=1,\dots,n,$ and $A \ge 0$ are integer.

Below we also consider the {\em One-Dimensional Bin Packing
Problem}. Given an integer number~$A$ (size of a bin) and
$k$~integer numbers $a_1,\dots,a_k$ (sizes of items), $a_i\le A, \
i=1,\dots,k$ it is required to locate the items into the minimal
number of bins, so that the sum of sizes of items in each bin does
not exceed~$A$.

The One-Dimensional Bin Packing Problem may be formulated as a
Boolean Linear Programming Problem the following way (a more
"standard" integer linear programming formulation can be found
e.g. in~\cite{MM}). Let a Boolean variable~$y_j$ be the indicator
of usage of a bin~$j, \ j=1,\dots,k$ and a Boolean
variable~$x_{ij}$ be the indicator of packing item~$i$ in bin~$j$,
$i,j=1,\dots,k$. Find
\begin{equation}\label{eq_bpack1}
\mbox{min} \sum_{j=1}^{k} y_j
\end{equation}
s. t.
\begin{equation}\label{eq_bpack3}
\sum_{j=1}^k x_{ij} = 1, \ \ i=1,\dots,k,
\end{equation}

\begin{equation}\label{eq_bpack8}
\sum_{i=1}^k a_i x_{ij} \le A, \ \ j=1,2,\dots,k,
\end{equation}

\begin{equation}\label{eq_bpack9}
y_j \ge x_{ij}, \ \ i =1,\dots,k, \ j =1,\dots,k,
\end{equation}

\begin{equation}\label{eq13_bpacka}
x_{ij} , \ y_j \in \{0,1\},\ \ i=1,\dots,k, \ j=1,2,\dots,k.
\end{equation}

Note that for solutions encoding it suffices to store only the
{\em matrix of assignments}~$(x_{ij})$, \since the vector
$(y_1,\dots,y_k)$ corresponding to such a matrix is uniquely
defined. Below we assume that $(k\times k)$-matrices of
assignments are used to encode the feasible
solutions and $n=k^2$.\\

The following special case of the well-known {Partition}
Problem~\cite{GJ} will be called {\em Bounded Partition}: Given
$2m$ positive integer numbers $\alpha_1,\ldots,\alpha_{2m}$, which
satisfy
\begin{equation}\label{eq:partition_limited}
\frac{B}{m+1}< \alpha_j<\frac{B}{m-1}, \ \ j=1,\dots,2m,
\end{equation}
where $B=\sum_{j=1}^{2m} \alpha_j/2$, {\em is there} a vector
$\x\in\{0,1\}^{2m}$, such that
\begin{equation}\label{eq:partition_usual}
\sum_{j=1}^{2m}\alpha_j x_j=B?
\end{equation}
The next lemma is due to P.~Schuurman and G.~Woeginger.

\begin{lemma} \label{lemm:spec_subsetsum}~{\rm \cite{SW}}
The Bounded Partition Problem is NP-complete.
\end{lemma}

{\bf Proof.} NP-completeness of this problem may be established
via reduction from the following NP-complete modification of
Partition Problem~\cite{GJ}: given a set of~$2m$ positive
integers~$\alpha'_j, \ j=1,\dots,2m$, it is required to recognize
existence of such~$\x\in\{0,1\}^{2m}$, that
\begin{equation}\label{eq:partition_two}
\sum_{j=1}^{2m} x_j=m \ \ \mbox{and}  \ \ \sum_{j=1}^{2m}
\alpha'_j x_j=\frac{1}{2} \sum_{j=1}^{2m} \alpha'_j.
\end{equation}
The reduction consists in setting $\alpha_i=\alpha'_i+M, \
i=1,\dots,2m,$ with a sufficiently large integer~$M$, e.g., $M =
2m \cdot \max\{\alpha'_j : j=1,\dots,2m\}$. Satisfaction
of~(\ref{eq:partition_limited}), as well as equivalence
of~(\ref{eq:partition_two}) and~(\ref{eq:partition_usual}), given
this set of parameters~$\{\alpha_i\}$, is verified
straightforwardly. $\Box$

\begin{theorem} \label{thm:BIN_KNAP_ORP} {\rm \cite{DE12}}
The ORPs for the One-Dimensional Knapsack Problem~(\ref{knap}) and
the One-Dimensional Bin-Packing Problem~(\ref{eq_bpack1}) --
(\ref{eq13_bpacka}) are NP-hard.
\end{theorem}

{\bf Proof.} 1. Consider ORP for Knapsack Problem~(\ref{knap}).
The NP-hardness of this problem can be established using a
polynomial-time Turing reduction of Bounded Partition Problem to
it. W. l. o. g. let us assume $m>2$.

Note that if an instance of Bounded Partition Problem has the
answer~"yes", then there exists a vector~$\x' \in \{0,1\}^{2m}$,
such that $\sum_{i=1}^{2m} \alpha_i x'_i = B$, and \since
$B/(m+1)<\alpha_i<B/(m-1), \ i=1,\dots,2m$, this vector contains
exactly~$m$ ones, which is less than~${2m-2}$ because~$m>2$. On
the contrary, if the instance of Partition Problem has the
answer~"no", then such a vector does not exist.

The Turing reduction of Bounded Partition Problem to the ORP for
One-Dimensional Knapsack problem is based on enumeration of
polynomial number of different pairs of parent solutions (and the
corresponding ORP instances). Assume $n=2m,$ $A=B$ and
${c_j=a_j=\alpha_j}, \ j=1,\dots,n,$ and enumerate all of the $2m
\choose 2$ pairs of variables with indices~$\{i_{\ell},
j_{\ell}\}, \ {\ell}=1,\dots, {2m \choose 2}$. For each
pair~$i_{\ell}, j_{\ell}$ we set $p^1_{i_{\ell}}=p^2_{j_{\ell}}=0$
and fill the remaining positions~$j\not\in \{i_{\ell}, j_{\ell}\}$
so that $p^1_j+p^2_j=1$ and each of the parent solutions contains
in total~$m-1$ ones (such parent solutions will be feasible \since
$a_j<A/(m-1), \ j=1,\dots,n$). The greatest value among the optima
of the constructed ORPs equals~$A$ iff the answer to the instance
of Partition Problem is "yes". This implies NP-hardness of the ORP
for One-Dimensional Knapsack Problem.


2. The proof of NP-hardness of the ORP for One-Dimensional
Bin-Packing Problem is based on a similar (but more time
demanding) Turing reduction from Bounded Partition Problem. Now we
assume $k=2m,$ $A=B$, and ${a_i=\alpha_i}, \ i=1,\dots,k$.
In what follows it is supposed that $m>4$.




Given an instance of Bounded Partition Problem, we enumerate a
polynomial number of parent solutions, choosing them in such a way
that (i)~$2m-4$ items in the offspring solution are packed into
the first two containers, (ii)~among them, a pair of "selected"
items may be packed only in bin~2, (iii)~four other "selected"
items may be packed either in bin~1 or in bin~3 optionally. Let us
describe this reduction in detail.

As in the first part of the proof we enumerate all of the $2m
\choose 2$ pairs of items~$\{i_{\ell}, i'_{\ell}\}, \
{\ell}=1,\dots, {2m \choose 2},$ aiming to fix the corresponding
variables~$\{x_{i_{\ell},1}, x_{i'_{\ell},1}\}$ to zero value.

For each of the pairs~$\{i_{\ell}, i'_{\ell}\}$ enumerate all
$2m-2 \choose 2$ pairs $\{u_{r}, u'_{r}\}, \ {r}=1,\dots, {2m-2
\choose 2}$ drawn from the rest of the items. Given
$\{i_{\ell},i'_{\ell}\}$ and $\{u_{r},u'_{r}\}$, enumerate all
$2m-4 \choose 2$ pairs $\{v_{s}, v'_{s}\}, \ {s}=1,\dots, {2m-4
\choose 2},$ in the rest of the items.

To ensure that for given $\ell,r$ and $s$, the items
$\{i_{\ell},i'_{\ell}\}$ in the offspring solution are packed in
bin~2, while items $u_{r}, u'_{r}, v_{s}, v'_{s}$ may be packed
only in bin~1 or bin~3, the pair of parent solutions ${\bf
p}^1=(p^1_{ij})$ and ${\bf p}^2=(p^2_{ij})$ is defined the
following way.

In the first column of parent solutions
$$
p^1_{i_{\ell},1}=p^1_{i'_{\ell},1}=p^2_{i_{\ell},1}=p^2_{i'_{\ell},1}=0,
$$
$$
p^1_{u_{r},1}=p^1_{u'_{r},1}=1, \ p^2_{u_{r},1}=p^2_{u'_{r},1}=0,
$$
$$
p^1_{v_{s},1}=p^1_{v'_{s},1}=0, \ p^2_{v_{s},1}=p^2_{v'_{s},1}=1
$$
and fill the remaining positions~$i\not\in \{i_{\ell},
i'_{\ell},u_{r}, u'_{r},v_{s}, v'_{s}\}$ so that
$p^1_{i,1}+p^2_{i,1}=1$ holds and each of the parent solutions
has~$m-1$ ones in column~1. These parent solutions satisfy
condition~(\ref{eq_bpack8}) for bin~$j=1$, \since ${a_i<A/(m-1),}
\ i=1,\dots,k$.

Let the second column in each of the parent solutions be identical
to the first column of the other parent, except for the components
corresponding to the six items mentioned above. Two entries~1 in
rows~$v_{s}$ and $v'_{s}$ of the parent solution~${\bf p}^1$ are
placed into column~$j=3$, rather than column~$j=2$. Two entries~1
in rows~$u_{r}$ and $u'_{r}$ of the parent solution~${\bf p}^2$
are placed into column~$j=3$, rather than column~$j=2$. Besides
that, in column~$j=2$ of both parent solutions the entries~1 are
placed in rows~$i_{\ell}$ and $i'_{\ell}$.

In each parent solution the second column contains~$m-1$
entries~1, thus condition~(\ref{eq_bpack8}) for bin~$j=2$ is
satisfied as well as in the case of~$j=1$. For bin~$j=3$ this
condition holds, \since ${a_i<A/4,} \ i=1,\dots,k$ when $m>4$.
Note that all feasible solutions to the ORP corresponding to a
triple of indices~$\ell, r, s$ contain the items~$i_{\ell},
i'_{\ell}$ in the second bin, while items $u_{r}$, $u'_{r}$,
$v_{s}$ and $v'_{s}$ may appear either in bin~1 or in bin~3.

If an instance of Bounded Partition Problem has the answer "yes"
then at least one of the constructed ORPs has the optimum
objective function value~$2$. Indeed, in such a case the
vector~$\x'$ that satisfies condition~(\ref{eq:partition_usual})
should have two entries $x'_{\hat{i}}=x'_{\bar{i}}=0$ for some
$\hat{i}, \bar{i}$. Besides that, there are four indices $\hat{u},
\bar{u}, \hat{v}$ and $\bar{v}$, such that
$x'_{\hat{u}}=x'_{\bar{u}}=x'_{\hat{v}}=x'_{\bar{v}}=1$, \since
this vector contains not less than~$m$ entries~1. The
corresponding ORP with $\{i_{\ell},i'_{\ell}\}=\{\hat{i},
\bar{i}\}$, $\{u_{r},u'_{r}\} =\{\hat{u}, \bar{u}\}$ and
$\{v_{s},v'_{s}\}=\{\hat{v},\bar{v}\}$ has a feasible
solution~$(x'_{ij})$, where the first column is identical
to~$\x'$, the entries of the second column are $x'_{i2}=1-x'_i, \
i=1,\dots,k$, and the rest of the columns are filled with zeros.

Conversely, if an optimal solution~$x^*_{ij}$ to one of the
constructed ORPs has the value~2, then setting $x_i=x^*_{i1}, \
i=1,\dots, k,$ we obtain equality~(\ref{eq:partition_usual}).
$\Box$\\


The One-Dimensional Bin Packing problem is contained as a special
case in a number of packing and scheduling problems, so the latter
theorem may be applicable in analysis of complexity of the ORPs
for these problems. In particular, Theorem~\ref{thm:BIN_KNAP_ORP}
implies NP-hardness of the ORP for the Transfer Line Balancing
Problem~\cite{DEG10}.

\subsection{Set Covering and Location Problems}
The next example of an NP-hard ORP is that for the Set Covering
Problem, which may be considered as a special case
of~(\ref{goal})-(\ref{bools}):
\begin{equation}\label{setcover}
\min\left\{{\bf cx} : {\bf Ax} \ge {\bf e}, \ \x \in
\{0,1\}^n\right\},
\end{equation}
where ${\bf A}$ is a Boolean $(m\times n)$-matrix; ${\bf
c}=(c_1,\dots,c_n);$ $c_j\ge 0, \ j=1,\dots,n$. Let us assume the
binary representation of solutions by the vector~$\x$. Given an
instance of the Set Covering Problem, one may construct a new
instance with a doubled set of columns in the matrix ${\bf
A}'=({\bf AA})$ and a doubled vector ${\bf
c}'={(c_1,\dots,c_n,c_1,\dots,c_n)}$. Then an instance of the
NP-hard Set Covering Problem~(\ref{setcover}) is equivalent to the
ORP for the modified set covering instance where the input
consists of $(m\times 2n)$-matrix ${\bf A}'$, $2n$-vector~${\bf
c}'$ and the feasible parent solutions $\p^1,\p^2,$ with $p^1_j=1,
p^2_{j}=0$ for $j=1,\dots,n$ and $p^1_j=0, p^2_{j}=1$ for
$j=n+1,\dots,2n$. So, the ORP for the Set Covering Problem is also
NP-hard.

Interestingly, in some cases the ORP may be even harder than the
original problem (assuming P$\ne$ NP). This can be illustrated on
the example of the Set Covering Problem. A special case of this
problem, defined by the restriction $a_{i,1}=1, i=1,\dots,m; \
c_1=0$ is trivially solvable: $\x=(1,0,0,\dots,0)$ is the optimal
solution. However, in the case $p^1_1=p^2_1=0$, the ORP becomes
NP-hard under this restriction.

The Set Covering Problem may be efficiently transformed to the
Simple Plant Location
Problem~(\ref{SPPLgoal0})-(\ref{SPPL_nonempty}) -- see e.g.
transformation~T3 \ in~\cite{KP83}. In this case the dimensions
$m$ and $n$ in both problems are equal, $C_i=c_i$ for
$i=1,\dots,n$ and
$$c_{ij}= \left\{
\begin{array}{cl}
 \sum_{k=1}^n c_k+1, & \mbox{ if } \ a_{ij}=0,\\
 0, & \mbox{ if } \ a_{ij}=1,
\end{array} \right. \ \ \mbox{for all} \ \ i=1,\dots,n, \ j=1,\dots,m.
$$
It is easy to verify that a vector~${\bf x}^*$ in the optimal
solution to this instance of the Simple Plant Location Problem
will be an optimal set covering solution as well. Thus, if the
solution representation in the Simple Plant Location Problem is
given only by the vector~$\bf x$, then this reduction meets the
conditions of Proposition~\ref{reduction0}. The subset of
solutions to the Simple Plant Location Problem~$\beta({\rm
Sol}_1(I))$  is characterized in this case by the threshold on
objective function $f_{\rm splp}({\bf y})<\sum_{j=1}^n c_j+1$,
which ensures that all constraints of the Set Covering Problem are
met. Therefore, an NP-hard ORP problem is efficiently reduced to
the ORP for~(\ref{SPPLgoal0})-(\ref{SPPL_nonempty}) and the
following proposition holds.

\begin{proposition}\label{SPLP_NP} {\rm \cite{ErECJ08}}
The ORP for the Simple Plant Location
Problem~(\ref{SPPLgoal0}),~(\ref{SPPL_nonempty}) is NP-hard.
\end{proposition}

The well-known $p$-Median Problem may be defined as a modification
of the Simple Plant Location
Problem~(\ref{SPPLgoal0}),~(\ref{SPPL_nonempty}): it suffices to
assume $C_k=0, \ j=1,\dots,n,$ and to substitude the
inequality~(\ref{SPPL_nonempty}) by constraint
\begin{equation}
\label{pmedian}
\sum_{i=1}^n x_i=p,
\end{equation}
where $1\le p \le n$ is a parameter given in the problem input.

\begin{proposition}\label{PM_NP} {\rm \cite{ErECJ08}}
The ORP for the $p$-Median
Problem~(\ref{SPPLgoal0}),~(\ref{pmedian}) is NP-hard.
\end{proposition}

{\bf Proof.} E.~Alexeeva, Yu.~Kochetov and A.~Plyasunov
in~\cite{AKP} propose a reduction of an NP-hard Graph Partitioning
Problem to the $p$-Median Problem with~$n=|V|$ and $p=|V|/2$,
where $V$ is the set of the graph vertices and $|V|$ is even.
Thus, this special case of the $p$-Median Problem is NP-hard as
well. Consider an ORP for this case of the $p$-Median Problem with
parent solutions $\p^1=(1,\dots,1,0,\dots,0)$ and
$\p^2=(0,\dots,0,1,\dots,1)$ of $n/2$ ones. Obviously, such ORP is
equivalent to the original $p$-Median Problem. \ $\Box$

\section{Travelling
Salesman Problem \label{sec:TSP}}

In this section we consider the Travelling Salesman Problem~(TSP):
suppose a digraph~$G$ without loops or multiple arcs is given. The
set of vertices of~$G$ is~$V$ and a set of arcs is~$A$. A weight
(length)~$c_{ij}\ge 0$ of each arc~$(i,j)\in A$ is given as well.
It is required to find a Hamiltonian circuit of minimum length.

If for each arc~$(i,j)\in A$ there exists a reverse one~${(j,i)\in
A}$ and $c_{ij}=c_{ji}$, then the TSP is called {\em symmetric}
and~$G$ is assumed to be an ordinary graph. Without this
assumption we will call the problem the {\em general case} of TSP.

Feasible solution to the TSP  may be encoded as a sequence of the
vertex numbers in the TSP tour, or as a permutation matrix where
the element in row~$i$ and column~$j$ equals one iff the
vertex~$j$ immediately follows the vertex~$i$ in the TSP tour.
(For the sake of consistency with Definition~\ref{def:NPO} one may
assume that the elements of the matrix are written sequentially in
a string~$\x \in \mbox{\rm Sol}$.)

Unfortunately there are~$|V|$ different sequences of vertices
encoding the same Hamiltonian circuit. The second encoding has an
advantage that a Hamiltonian circuit is uniquely represented by a
permutation matrix. Therefore in what follows we assume the second
encoding. If this encoding is used in the symmetric case, it is
sufficient to define only the elements above the diagonal of the
matrix, so the rest of the elements are dismissed from subsequent
consideration in the symmetric case.

The encoding by permutation matrix defines an ORP that consists in
finding a shortest travelling salesman's tour which coincides with
two given feasible parent solutions in those arcs (or edges) which
belong to both parent tours and does not contain the arcs (or
edges) which are absent in both parent solutions.


\subsection{Symmetric Case} \label{subsec:NP-hard_sym}
In~\cite{IPS82} it is proven that recognition of Hamiltonian grid
graphs (the {\em Hamilton Cycle Problem}) is NP-complete. Recall
that a graph~$G'=(V',E')$ with vertex set~$V'$ and edge set~$E'$
is called a {\em grid} graph, if its vertices are the integer
vectors $v=(x_v,y_v)\in {\bf Z}^2$ on plane, i.e., $V' \subset
{\bf Z}^2$, and a pair of vertices is connected by an edge iff the
Euclidean distance between them is equal to~1. Here and below,
$\bf Z$ denotes the set of integer numbers. Let us call the edges
that connect two vertices in~${\bf Z}^2$ with equal first
coordinates {\em vertical edges}. The edges that connect two
vertices in~${\bf Z}^2$ with equal second coordinates will be
called {\em horizontal edges}.

Let us assume that~$V'>4$, graph~$G'$ is connected and there are
no bridges in $G'$ (note that if any of these assumptions is
violated, then existence of a Hamiltonian cycle in~$G'$ can be
recognized in polynomial time). Now we will construct a reduction
from the Hamilton Cycle Problem for~$G'$ to an ORP for a complete
edge-weighted graph~$G=(V,E)$, where $V=V'$.

\begin{figure}
\begin{center}
\vspace{2em}
\includegraphics[height=10cm,width=10cm]{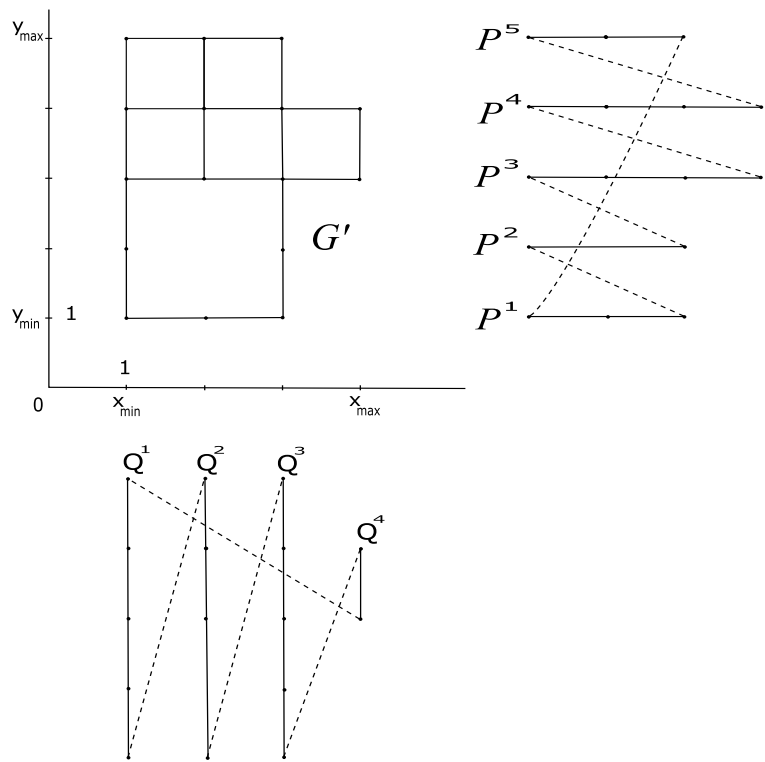}\\
\caption{Example of two parent tours used in reduction from
Hamilton Cycle Problem to ORP in symmetric case.}
\label{fig:grid_reduct}
\end{center}
\end{figure}

Let the edge weights~$c_{ij}$ in~$G$ be defined so that if a pair
of vertices $\{v_i,v_j\}$ is connected by an edge of~$G'$,
then~$c_{ij}=0$; all other edges in~$G$ have the weight~1.
Consider the following two parent solutions of the TSP on
graph~$G$ (an example of graph~$G'$ and two parent solutions for
the corresponding TSP is given in Fig.~\ref{fig:grid_reduct}).

Let $y_{\min}=\min_{v \in V'} y_v, \ y_{\max}=\max_{v \in V'}
y_v.$ For any integer $y \in \{y_{\min},\dots,$ $y_{\max}\}$, the
horizontal chain that passes through vertices $v\in V'$ with
$y_v=y$ by increasing values of coordinate~$x$ is denoted
by~$P^{y}$. Let the first parent tour follow the chains $
P^{y_{\min}},P^{y_{\min}+1},\dots,P^{y_{\max}}$, connecting the
right-hand end of each chain~$P^y$ with $y<y_{\max}$ to the
left-hand end of the chain~$P^{y+1}$. Note that these connections
never coincide with vertical edges because~$G'$ has no bridges. To
create a cycle, connect the right-hand end~$v_{\rm TR}$ of the
chain~$P^{y_{\max}}$ to the left-hand end~$v_{\rm BL}$ of the
chain~$P^{y_{\min}}$.

The second parent tour is constructed similarly using the vertical
chains. Let $x_{\min}=\min_{v \in V'} x_v$, $x_{\max}=\max_{v \in
V'} x_v$. For any integer $x \in \{x_{\min},\dots,x_{\max}\},$ the
vertical chain that passes monotonically in~$y$ through the
vertices $v\in V'$, such that $x_v=x$, is denoted by~$Q^{x}$. The
second parent tour follows the chains
$Q^{x_{\min}},Q^{x_{\min}+1},\dots,Q^{x_{\max}}$, connecting the
lower end of each chain~$Q^x$ with $x<x_{\max}$ to the upper end
of chain~$Q^{x+1}$. These connections never coincide with
horizontal edges since $G'$ has no bridges. Finally, the lower
end~$v_{\rm RB}$ of chain~$Q^{x_{\max}}$ is connected to the upper
end~$v_{\rm LT}$ of chain~$Q^{x_{\min}}$.

Note that the constructed parent tours have no common edges.
Indeed, common slanting edges do not exist since $V'>4$. The
horizontal edges belong to the first tour only, except for the
situation where $y_{v_{\rm RB}}=y_{v_{\rm LT}}$ and the edge
$\{v_{\rm RB},v_{\rm LT}\}$ of the second tour is oriented
horizontally. But if the first parent tour included the edge
$\{v_{\rm RB},v_{\rm LT}\}$ in this situation, then the edge
$\{v_{\rm RB},v_{\rm LT}\}$ would be a bridge in graph~$G'$.
Therefore the parent tours can not have the common horizontal
edges. Similarly the vertical edges belong to the second tour
only, except for the case where $x_{v_{\rm TR}}=x_{v_{\rm BL}}$
and the edge $\{v_{\rm TR},v_{\rm BL}\}$ of the first tour is
oriented vertically. But in this case the parent tour can not
contain the edge $\{v_{\rm TR},v_{\rm BL}\}$, since $G'$ has no
bridges.

Note also that the union of edges of parent solutions
contains~$E'$. Consequently, any Hamiltonian cycle in graph~$G'$
is a feasible solution of the ORP. At the same time, a feasible
solution of the ORP has zero value of objective function iff it
contains only the edges of~$E'$. Therefore, the optimal value of
objective function in the ORP under consideration is equal to~0
iff there exists a Hamiltonian cycle in graph~$G'$. So, the
following theorem is proven.

\begin{theorem} \label{th_cyc_NP_hard} {\rm \cite{Er_EvoCOP11}}
Optimal recombination for the TSP in the symmetric case is
strongly NP-hard.
\end{theorem}

In~\cite{IPS82} it is also proven that recognition of grid graphs
with a {\em Hamiltonian path} is NP-complete. Optimal
recombination for this problem consists in finding a shortest
Hamiltonian path, which uses those edges where both parent tours
coincide, and does not use the edges absent in both parent tours.
The following theorem is proved analogously to
Theorem~\ref{th_cyc_NP_hard}.

\begin{theorem} \label{th_path_NP_hard} {\rm \cite{Er_EvoCOP11}}
Optimal recombination for the problem of finding the shortest
Ham\-il\-ton\-ian path in a graph with arbitrary edge lengths is
strongly NP-hard.
\end{theorem}

Note that in the proof of Theorem~\ref{th_path_NP_hard}, unlike in
Theorem~\ref{th_cyc_NP_hard}, it is impossible simply to exclude
the cases where graph~$G'$ has bridges. Instead, the reduction
should treat separately each maximal (by inclusion) subgraph
without bridges.


Many scheduling problems with setup times contain the problem of
finding the shortest Ham\-il\-ton\-ian path in a digraph as a
special case. In this case the vertices correspond to jobs, the
arcs correspond to setups and the arc lengths define the setup
times. In view of numerous applications of scheduling problems
with setup times, in Section~\ref{sec:makespan} the problem of
finding the shortest Ham\-il\-ton\-ian path in a digraph is
treated as a scheduling problem.

\subsection{The General Case} \label{subsec:NP-hard_general}

In the general case of TSP the ORP is not a more general problem
than the ORP considered in Subsection~\ref{subsec:NP-hard_sym}
because in the problem input we have two directed parent paths,
while in the symmetric case the parent paths were undirected. Even
if the distance matrix~$(c_{ij})$ is symmetric, a pair of directed
parent tours defines a significantly different set of feasible
solutions, compared to the undirected case. Therefore, the general
case requires a separate consideration of ORP complexity.

\begin{theorem} \label{th_path_NP_hard_general} {\rm \cite{Er_EvoCOP11}}
Optimal recombination for the TSP in the general case is strongly
NP-hard.
\end{theorem}

{\bf Proof.} We use a modification of the textbook reduction of
the Vertex Cover Problem to the TSP~\cite{GJ}.

Suppose an instance of a Vertex Cover Problem is given as a
graph~$G'=(V',E')$. It is required to find a vertex cover of
minimal size in~$G'$. Let us assume that the vertices in~$V'$ are
enumerated, i.e. $V'=\{v_1,\dots,v_n\}$, where $n=|V'|$, and let
$m=|E'|$.

Consider a complete digraph~$G=(V,A)$ where the set of
vertices~$V$ consists of $|E'|$ {\em cover-testing} components,
each one containing 12 vertices: $V_e=\{(v_i,e,k),(v_j,e,k): 1 \le
k \le 6\}$ for each $e=\{v_i,v_j\} \in E', \ i<j$. Besides that,
$V$ contains $n$ {\em selector} vertices denoted by
$a_1,\dots,a_{n}$, and a {\em supplementary} vertex~$a_{n+1}$.

\begin{figure}
\begin{center}
\vspace{2em}
\includegraphics[height=8cm,width=10.5cm]{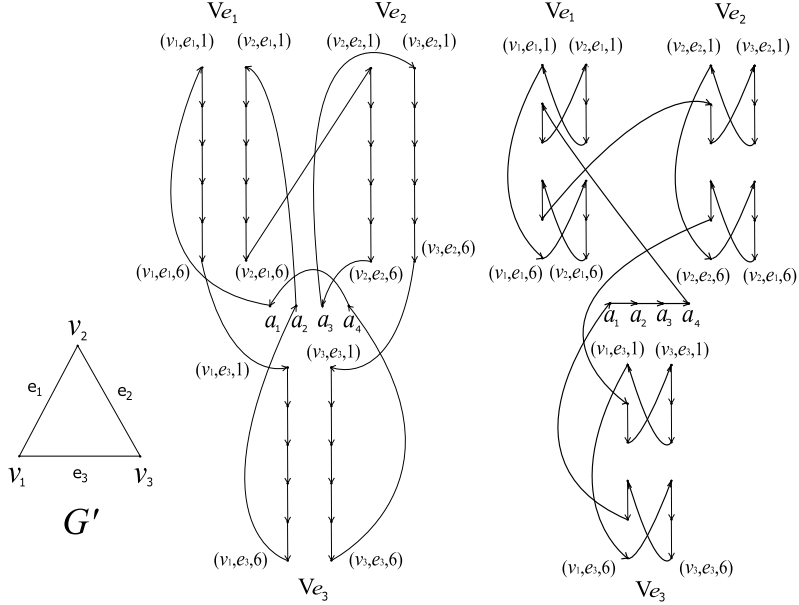}\\
\caption{A pair of parent circuits for the case of $G'=K_3$. It is
supposed that the incident edges are enumerated as follows. For
vertex~$v^1: \ e^{v_1,1} = e_1, \ e^{v_1,2}=e_3;$ for vertex~$v^2:
\ e^{v_2,1} =e_1, \ e^{v_2,2}=e_2;$ for vertex~$v^3: \
e^{v_3,1}=e_2, e^{v_3,2}=e_3$.}\label{fig:general_2parents}
\end{center}
\end{figure}

Let the parent tours in graph~$G$ be the two circuits defined
below (an example of a pair of such circuits for the case of
$G'=K_3$ is provided in Fig.~\ref{fig:general_2parents}).

1. Each cover-testing component~$V_e$, where $e=\{v_i,v_j\} \in
E'$ and $i < j$ is visited twice by the first tour. The first time
it visits the vertices that correspond to~$v_i$ in the sequence
\begin{equation} \label{eq:left_post}
(v_i,e,1), \dots, (v_i,e,6),
\end{equation}
the second time it visits the vertices corresponding to~$v_j$, in
the sequence
\begin{equation} \label{eq:right_post}
(v_j,e,1), \dots, (v_j,e,6).
\end{equation}

2. The second tour goes through each cover-testing
component~$V_e$, where $e=\{v_i,v_j\} \in E'$ and $i < j$ in the
following sequence:

$$
 (v_i,e,2), (v_i,e,3), (v_j,e,1), (v_j,e,2), (v_j,e,3), (v_i,e,1),
$$
$$
 (v_i,e,6), (v_j,e,4), (v_j,e,5), (v_j,e,6), (v_i,e,4),
 (v_i,e,5).
$$

The first parent tour connects the cover-testing components as
follows. For each vertex $v\in V'$ order arbitrarily the edges
incident to~$v$ in graph~$G'$ in sequence: $e^{v,1}, e^{v,2},
\dots, e^{v,deg(v)},$ where $deg(v)$ is the degree of vertex~$v$
in~$G'$. In the cover-testing components, following the chosen
sequence $e^{v,1}, e^{v,2}, \dots, e^{v,deg(v)}$, this tour passes
6 vertices in each of the components $(v,e,k), \ k=1,\dots,6, \ e
\in \{e^{v,1}, e^{v,2}, \dots, e^{v,deg(v)}\}$. Thus, each vertex
of any cover-testing component~$V_e$, $e=\{u,v\} \in E'$ will be
visited by one of the two 6-vertex sub-tours.

The second tour passes the cover-testing components in an
arbitrary order of edges $V_{e_1},\dots,V_{e_{m}}$, entering each
component $V_{e_{k}}$ for any $e_k=\{v_{i_k},v_{j_k}\}\in E', \
i_k<j_k, \ k=1,\dots,m,$ via vertex $(v_{i_k},e_k,2)$ and exiting
through vertex $(v_{i_k},e_k,5)$. Thus, a sequence of vertex
indices $i_1,\dots,i_m$ is induced (repetitions are possible). In
what follows, we will need the beginning~$i_1$ and the end~$i_m$
of this sequence.

The parent sub-tours described above are connected to form two
Hamiltonian circuits in~$G$ using the vertices
$a_1,\dots,a_{n+1}$. The first circuit is completed using the arcs
$$
 \Big(a_1,(v_1,e^{v_1,1},1)\Big), \
 \Big((v_1,e^{v_1,deg(v_1)},6),a_2\Big), \
$$
$$
 \Big(a_{2},(v_{2},e^{v_{2},1},1)\Big), \
 \Big((v_{2},e^{v_{2},deg(v_{2})},6),a_{3}\Big),
$$
$$
  \dots,
$$
$$
 \Big(a_{n},(v_{n},e^{v_{n},1},1)\Big), \
 \Big((v_{n},e^{v_{n},deg(v_{n})},6),a_{n+1}\Big),
 \Big(a_{n+1},a_1\Big).
$$
The second circuit is completed by the arcs
$$
 \Big(a_1,a_2\Big), \dots, \Big(a_{n-1},a_n\Big), \Big(a_{n},a_{n+1}\Big),
$$
$$
 \Big(a_{n+1},(v_{i_1},e_1,2)\Big), \ \Big((v_{i_m},e_m,5), a_{1}\Big).
$$

Assign unit weights to all arcs $\Big(a_i,(v_i,e^{v_i,1},1)\Big),
\ i=1,\dots,n$ in the complete digraph~$G$. Besides that, assign
weight~$n+1$ to all arcs of the second tour which are connecting
the components $V_{e_1},\dots,V_{e_{m}}$, the same weights are
assigned to the arcs $\Big(a_{n+1},(v_{i_1},e_1,2)\Big)$ and
$\Big((v_{i_m},e_m,5), a_{1}\Big)$. All other arcs in~$G$ are
given weight~0.

\begin{figure}
\begin{center}
\vspace{2em}
\includegraphics[height=8cm,width=5cm]{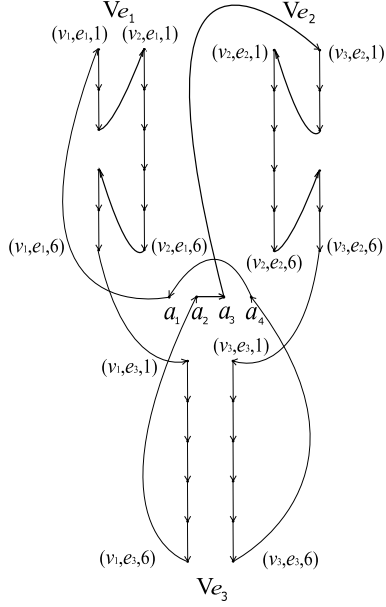}\\
\caption{An ORP solution $R(C)$ corresponding to the vertex cover
$\{v_1,v_3\}$ of graph $G'=K_3$.}\label{fig:general_child}
\end{center}
\end{figure}

Note that for any vertex cover~$C$ of graph~$G'$, the set of
feasible solutions of ORP with two parents defined above contains
a circuit~$R(C)$ with the following structure (see an example of
such a circuit for the case of $G'=K_3$ in
Fig.~\ref{fig:general_child}).

For each $v_i \in C$ the circuit~$R(C)$ contains the arcs
$\Big(a_i,(v_i,e^{v_i,1},1)\Big)$ and
$\Big((v_i,e^{v_i,deg(v_i)},6), a_{i+1}\Big)$. The
components~$V_e, \ e \in \{e^{v_i,1}, e^{v_i,2}, \dots,
e^{v_i,deg(v_i)}\}$ are connected together by the arcs from the
first tour. For each vertex~$v_i$ which does not belong to~$C$,
the circuit~$R(C)$ has an arc $(a_i,a_{i+1})$. Also, $R(C)$ passes
the arc $(a_{n+1},a_1)$.

The circuit~$R(C)$ visits each cover-testing component~$V_e$ by
one of the two ways:

1. If both endpoints of an edge~$e$ belong to~$C$, then~$R(C)$
passes the component following the same arcs as the first parent
tour.

2. If~$e=\{u,v \}$, $u \in C$, $v \not\in C$, then~$R(C)$ visits
the vertices of the component in sequence
$$
 (u,e,1), (u,e,2), (u,e,3), \ \
 (v,e,1), \dots, (v,e,6), \ \
 (u,e,4), (u,e,5), (u,e,6).
$$
One can check straightforwardly that this sequence does not
violate the ORP constraints.

In general, the circuit~$R(C)$ is a feasible solution to the ORP
because, on one hand, all arcs used in~$R(C)$ are present at least
in one of the parent tours. On the other hand, both parent tours
contain only the arcs of the type
$$
 \Big((u,e,2),(u,e,3)\Big), \
 \Big((u,e,4),(u,e,5)\Big), \
 \Big((v,e,1),(v,e,2)\Big),
$$
$$
 \Big((v,e,2),(v,e,3)\Big), \
 \Big((v,e,4),(v,e,5)\Big),
 \Big((v,e,5),(v,e,6)\Big)
$$
within the cover-testing components~$V_e$, $e=\{u,v\} \in E'$,
where vertex~$u$ has a smaller index than~$v$. All of these arcs
belong to~$R(C)$. The total weight of circuit~$R(C)$ is~$|C|$.

Now each feasible solution~$R$ to the constructed ORP defines a
set of vertices~$C(R)$ as follows: $v_i, \ i\in\{1,\dots,n\}$
belongs to~$C(R)$ iff $R$ contains an arc
$\Big(a_i,(v_i,e^{v_i,1},1)\Big)$.

Let us consider only such ORP solutions~$R$ that have the
objective value at most~$n$. These solutions do not contain the
arcs that connect the cover-testing components in the second
parent tour. They also do not contain the arcs
$\Big(a_{n+1},(v_{i_1},e_1,2)\Big)$ and $\Big((v_{i_m},e_m,5),
a_{1}\Big)$. Note that the set of such ORP solutions is non-empty,
e.g. the first parent tour belongs to it.

Consider the case where the arc $\Big(a_i,(v_i,e^{v_i,1},1)\Big)$
belongs to~$R$. Each cover-testing component~$V_e$ with
$e=\{v_i,v_j\} \in E'$ in this case may be visited in one of the
two possible ways: either the same way as in the first parent tour
(in this case, $v_j$ must also be chosen into~$C(R)$ since~$R$ is
Hamiltonian), or in the sequence
$$
 (v_i,e,1), (v_i,e,2), (v_i,e,3), \ \
 (v_j,e,1), \dots, (v_j,e,6), \ \
 (v_i,e,4), (v_i,e,5), (v_i,e,6)
$$
(in this case, $v_j$ will not be chosen into~$C(R)$). In view of
the assumption that the arc $\Big(a_i,(v_i,e^{v_i,1},1)\Big)$
belongs to~$R$, the cover-testing components~$V_e, \ e \in
\{e^{v_i,1}, e^{v_i2}, \dots, e^{v_i,deg(v_i)}\}$ are connected by
the arcs of the first tour, and besides that, $R$ contains the arc
$\Big((v_i,e^{v_i,deg(v_i)},6), a_{i+1}\Big)$. Note that the total
length of the arcs in~$R$ equals~$|C(R)|$, and the set~$C(R)$ is a
vertex cover in graph~$G'$, because the tour~$R$ passes each
component~$V_e$ in a way that guarantees coverage of each
edge~$e\in E'$.

To sum up, there exists a bijection between the set of vertex
covers in graph~$G'$ and the set of feasible solutions to the ORP
of length at most~$n$. The values of objective functions are not
changed under this bijection, therefore the statement of the
theorem follows. $\Box$

\subsection{Transformation of the ORP into TSP on Graphs With
Bounded Vertex Degree} \label{subsec:alg}

In this Subsection, the ORP problems are connected to the TSP on
graphs (digraphs) with bounded vertex degree, arbitrary positive
edge (arc) weights and a given set of {\em forced} edges (arcs).
It is required to find a shortest Hamiltonian cycle (circuit) in
the given graph (digraph) that passes all forced edges (arcs).

\subsubsection{General Case} \label{subsubsec:Sol_gen}

Consider the general case of ORP for the TSP, where we are given
two parent tours~$A_1, A_2$  in a complete digraph $G=(V,A)$. This
ORP problem may be transformed into the problem of finding a
shortest Hamiltonian circuit in a supplementary
digraph~$G'=(V',A')$. The digraph~$G'$ is constructed on the basis
of~$G$ by excluding the set of arcs $A\backslash (A_1 \cup A_2)$
and contracting each path that belongs to both parent tours into a
pseudo-arc of the same length and the same direction as those of
the path. The lengths of all other arcs that remained in~$G'$ are
the same as they were in~$G$. A shortest Hamiltonian circuit~$C'$
in~$G'$ transforms into an optimum of the ORP problem by
substitution of each pseudo-arc in~$C'$ with the path that
corresponds to it.

Note that there are at most two ingoing arcs and at most two
outgoing arcs for each vertex in~$G'$. The TSP on such a digraph
is equivalent to the TSP on a cubic digraph~$G''=(V'',A'')$, where
each vertex~$v\in V'$ is substituted by two vertices
$\check{v},\hat{v}$, connected by an {\em artificial} arc
$(\check{v},\hat{v})$ of zero length. All arcs that entered~$v$,
now enter~$\check{v}$, and all arcs that left~$v$ are now outgoing
from~$\hat{v}$. Let an arc~$e\in A''$ be forced, if it corresponds
to a pseudo-arc in~$G'$. Such arcs~$e\in A''$ are called
pseudo-arcs as well.

A solution to the TSP problem on digraph~$G''$ may be obtained
through enumeration of all feasible solutions to a TSP with forced
edges on a supplementary graph~$\bar{G}=(V'',\bar{E})$. Here, a
pair of vertices $u,v$ is connected iff these vertices were
connected by an arc (or a pair of arcs) in the digraph~$G''$. An
edge $\{u,v\}\in \bar{E}$ is assumed to be forced if $(u,v)$ or
$(v,u)$ is a pseudo-arc or an artificial arc in the digraph~$G''$.
A set of forced edges in~$\bar{G}$ will be denoted by~$\bar{F}$.
All Hamiltonian cycles in~$\bar{G}$ w.r.t. the set of forced edges
may be enumerated by means of the algorithm proposed
in~\cite{Eppst} in time $O(|V''| \cdot
2^{(|\bar{E}|-|\bar{F}|)/4})$. Then, for each Hamiltonian
cycle~$Q$ from~$\bar{G}$ in each of the two directions we can
check if it is possible to pass a circuit in~$G''$ through the
arcs corresponding to edges of~$Q$, and if possible, compute the
length of the circuit. This takes $O(|V''|)$ time for each
Hamiltonian cycle. Note that $|\bar{E}|-|\bar{F}| = d \le |E'| \le
2n$, where $d$ is the number of arcs which are present in one of
the parents only. Consequently, the time complexity of solving the
ORP on graph~$G$ is~$O(n \cdot 2^{d/4}),$ which is~$O(n \cdot
1.42^{n})$.

Implementation of the method described above may benefit in the
cases where the parent solutions have many arcs in common.

\subsubsection{Symmetric Case} \label{subsubsec:Sol_sym}

Suppose the symmetric case takes place and two parent Hamiltonian
cycles in graph $G=(V,E)$ are defined by two sets of edges $E_1$
and $E_2$. Let us construct a reduction of the ORP in this case to
a TSP with a set of forced edges on a graph where the vertex
degree is at most~4.

Similar to the general case, the ORP reduces to the TSP on a
graph~$G'=(V',E')$ obtained from~$G$ by exclusion of all edges
that belong to $E\backslash (E_1 \cup E_2)$ and contraction of all
paths that belong to both parent tours. Here, by contraction we
mean the following mapping. Let~$P_{uv}$ be a path with endpoints
in $u$ and $v$, such that the edges of $P_{uv}$ belong to $E_1
\cap E_2$ and $P_{uv}$ is not contained in any other path with
edges from $E_1 \cap E_2$. Assume that contraction of the
path~$P_{uv}$ maps all of its vertices and edges into one forced
edge $\{u,v\}$ of zero length. All other vertices and edges of the
graph remain unchanged. Let~$F'$ denote the set of forced edges
in~$G'$, which are introduced when the contraction is applied to
all paths wherever possible.

The vertex degrees in~$G'$ are at most~4, and $|V'| \le n$. If an
optimum of the TSP on graph~$G'$ with the set of forced edges~$F'$
is found, then substitution of all forced edges by the
corresponding paths yields an optimal solution to the ORP problem.
(Note that the objective functions of these two problems differ by
the total length of contracted paths.)

The search for an optimum to the TSP on graph~$G'$ may be carried
out by means of the randomized algorithm proposed in~\cite{Eppst}
for solving TSP with forced edges on graphs with vertex degree at
most~4. Besides the problem input data this algorithm is given a
value~$p$, which sets the desired probability of obtaining the
optimum. If $p \in [0,1)$ is a constant which does not depend on
the problem input, then the algorithm has time
complexity~$O((27/4)^{n/3})$, which is $O(1.89^n)$.

As it was noted in Subsection~\ref{subsec:GA}, when the crossover
operator is used in a GA, an additional parameter~$P_{\rm c}$ may
be defined to tune the probability of performing recombination. If
such a parameter is given, $P_{\rm c} \in [0,1),$ then~one may
assign $p=P_{\rm c}$. In case $P_{\rm c}=1$, the optimal
recombination may be performed using a deterministic modification
of the algorithm from~\cite{Eppst} (corresponding to~$p=1$) which
requires greater computation time.

There may be some room for improvement of the algorithms, proposed
in~\cite{Eppst} for the TSP on graphs with vertex degrees at most
3 or 4 and forced edges, in terms of the running time. Thus, it
seems to be important to continue studying this modification of
the TSP.

\section{Makespan Minimization on Single Machine}
\label{sec:makespan}

Consider the Makespan Minimization Problem on a Single Machine,
denoted by~$1|s_{vu}|C_{\max}$, which is equivalent to the problem
of finding the shortest Ham\-il\-ton\-ian path in a digraph.

The input consists of a set of jobs~${V}=\{v_1,\dots,v_k\}$ with
positive processing times ${p_v}$, ${v\in V}$. All jobs are
available for processing at time zero, and preemption is not
allowed. A sequence dependent setup time is required to switch a
machine from one job to another. Let $s_{vu}$ be the a
non-negative setup time from job~$v$ to job~$u$ for all ${v,u\in
V}$, where~${v\ne u}$. The goal is to schedule the jobs on a
single machine so as to minimize the maximum job completion time,
the so-called makespan~$C_{\max}$.

Let  $\pi=(\pi_1,\dots,\pi_{k})$ denote a permutation of the jobs,
\ie $\pi_i$ is the $i$-th job on the machine, ${i=1,\dots,k}$. Put
$s(\pi)=\sum_{i=1}^{k-1}s_{\pi_i,\pi_{i+1}}$. Then the
problem~$1|s_{vu}|C_{\max}$ is equivalent to finding a
permutation~$\pi^*$ that minimizes the total setup
time~$s(\pi^*)$.

We assume that the binary  encoding of solutions to this
NP~optimization problem is given by a permutation matrix, where
the element in row~$i$, column~$u$ equals~1 iff the $i$-th
executed job is the job~$u$. For the sake of convenience, however,
we will continue referring to feasible solutions in terms of
permutations where appropriate.



Note that the permutation matrices could be used for encoding the
solutions to problem~${1|s_{vu}|C_{\max}}$ so that a unit element
of the matrix reflects a setup between a pair of jobs (similar to
the encoding of TSP solutions in Section~\ref{sec:TSP}).
Experimental studies of GAs indicate, however, that the solution
encodings based on the sequence of jobs (as the one used in this
section) yield better results in solving the scheduling
problems~\cite{Reev}.

\subsection{NP-Hardness of Optimal Recombination}
\label{subsec:complexity} \sloppy
\par
{In what follows, we will use some remarkable results known for
the Shortest Hamiltonian Path Problem with Vertex Requisitions:}
given a complete  
digraph~$G=(X,U)$, where ${X=\{x_1,\dots,x_n\}}$ is the set of
vertices, $U=\{(x,y):\ x,y \in X, x\ne y\}$ is the set of arcs
with nonnegative weights ${\rho(x,y),}$ ${(x,y)\in U}$. Besides
that, a family of vertex subsets (requisitions) $X^i\subseteq X,\
i=1,\dots,n,$ is given, such that:\\
$\mathcal{C}1$:~$|X^i|\leqslant 2$ for all ${i=1,\dots,n}$;\\
$\mathcal{C}2$:~${1 \leqslant|\{i:\ x\in X^i,\ i=1,\dots,n\}|\leqslant 2}$ for all ${x\in X}$;\\
$\mathcal{C}3$:~if ${x\in X^i}$ and ${x\in X^j}$, where ${i\ne
j}$, then ${|X^i|=|X^j|=2}$, and if ${x\in X^i}$ for a unique~$i$,
then ${|X^i|=1}$.

Let $F$ denote the set of the bijections from~$X_n=\{1,\dots,n\}$
to~$X$ that satisfy the condition $f(i)\in X^i,\ i=1,\dots,n,$ for
all~$f\in F$. The problem asks for a mapping~${f}^*\in F$, such
that ${\rho}({f}^*)=\min\limits_{f\in F} {\rho}(f)$, where
${\rho}({f})=\sum\limits_{i=1}^{n-1} \rho({f}(i),{f}(i+1))$
for~$f\in F$. In what follows, this problem is denoted
by~${{\mathcal I}}$.

There always exists at least one  feasible solution~$f^1$ to
Problem~${{\mathcal I}}$. Indeed, such a solution exists iff there
is a perfect matching~$W$ in the bipartite
graph~$\bar{G}=(X_{n},X,\bar{U})$ where the subsets of vertices of
bipartition~$X_n,$ $X$ have equal size and the set of edges is
$\bar{U}=\{(i,x):$ ${i\in X_{n},\ x\in X^i}\}$. Note that if the
degree of a vertex~$i\in X_n$ in~$\bar{G}$ equals~$d$
(${1\leqslant d\leqslant 2}$) then, in view of
conditions~$\mathcal{C}2$ and $\mathcal{C}3$, the degree of all
vertices adjacent to~$i$ is also equal to~$d$. Thus for any
$Y\subseteq X_{n}$ holds~$|Y|\leqslant |\{x\in X:\ x\in X^i,\ i\in
Y\}|$ and the existence of~$W$ follows from the K\"{o}nig-Hall
Theorem~\cite{BK}. Besides that, the perfect matching
$W=\{(1,x^1),(2,x^2),\dots,(n,x^n)\}\subseteq \bar{U}$ may be
found in polynomial time using the K\"{o}nig-Hall
Algorithm~\cite{BK}. A feasible solution to problem~${{\mathcal
I}}$ is obtained assuming $f^1(i)=x^i,\ i=1,\dots,n$.

It is clear that with $|X^i|=1,\ i=1,\dots,n$, the
problem~${{\mathcal I}}$ is trivial, since the feasible solution
is unique. Therefore in what follows we shall assume that there
exists such~$i\in X_n$ that $|X^i|=2$. Then there is at least one
more feasible solution~$f^2$ to the problem~${{\mathcal I}}$,
where $f^2(i)= X^{i}\backslash \{f^1(i)\}$ for such~$i$ that
$|X^i|=2$, and $f^2(i)=f^1(i)$ otherwise.

Let us now proceed to complexity analysis of the ORP
for~$1|s_{vu}|C_{\max}$. First of all note that the
problem~${\mathcal I}$ reduces to it. Indeed, associate each
vertex~$x_i\in X$ of digraph~$G$ to a job~$v_i$, $i=1,\dots,n$,
let the number of jobs be~$n$ and let the setup
times~$s_{v_i,v_j}$ be equal to~$\rho(x_i,x_j)$ for all ${v_i,v_j
\in V}$, $i\ne j$. Assuming $\pi^1=f^1$ and $\pi^2=f^2$, we obtain
a polynomial-time reduction of problem~${\mathcal I}$ to the ORP
under consideration. In view of properties of this reduction,
if~${\mathcal I}$ were strongly NP-hard, this would imply that the
ORP for~$1|s_{vu}|C_{\max}$ is strongly NP-hard as well.

In~\cite{SAI}, A.I.~Serdyukov showed the strong NP-hardness of the
TSP with Vertex Requisitions, which is the TSP with a family of
requisitions defined as above, except that
conditions~$\mathcal{C}2$ and $\mathcal{C}3$ are dismissed, and
the goal is to find such a mapping~${\tilde{f}^*}$, that
${\tilde{\rho}(\tilde{f}^*)=\min\limits_{f\in F}
\tilde{\rho}(f)}$, where
$\tilde{\rho}({f})=\sum\limits_{i=1}^{n-1}
\rho({f}(i),{f}(i+1))+\rho({f}(n),{f}(1))$ for any~$f\in F$. Let
us denote this problem by~${\tilde{{\mathcal I}}}$. In what
follows it will be shown via a Turing reduction from
problem~${\tilde{{\mathcal I}}}$ that problem~${\mathcal I}$ is
NP-hard in the strong sense.

\begin{proposition} {\rm \cite{ErKo12}}
The problem~${\mathcal I}$ is strongly NP-hard. \label{prop1}
\end{proposition}

{\bf Proof.} Let us show that given an instance of
problem~$\tilde{{\mathcal I}}$ with a family of
requisitions~${X^i}$, ${i=1,\dots,n}$, it is possible to construct
efficiently an equivalent family of requisitions that will satisfy
conditions~$\mathcal{C}1$~--~$\mathcal{C}3$ or, alternatively, to
prove that the instance has no feasible solutions.

The equivalent family of requisitions is constructed by the
following sequence of transformations, where the vertices and
requisitions are labelled as {\em fathomed} or {\em unfathomed}.
Initially all vertices and requisitions are labelled as unfathomed.\\

{\bf 1.}~If there exists a vertex~$x\in X$ such that $\{i\in X_n:
\ x\in X^i\}=\emptyset$, then problem~$\tilde{{\mathcal I}}$
has no feasible solutions. No further transformations required.\\

{\bf 2.}~Perform the following operations until only the
two-element requisitions will remail among the unfathomed ones:
find an unfathomed subset~$X^i=\{x\}$ (\ie $|X^i|=1$) and delete
the vertex~$x$ from the other requisitions it belongs to; in case
the resulting family of requisitions contains such~$X^j$ that
$|X^j|=0$, this implies that~$\tilde{{\mathcal I}}$ has no
feasible solutions and no further transformations are required;
otherwise, label the vertex~$x$ and the subset~$X^i$
as fathomed.\\

{\bf 3.}~Perform the following operations until among the
unfathomed vertices there will be only the vertices that belong to
exactly~2 requisitions and each of these requisitions is of
cardinality~2: find an unfathomed vertex~$x$ that belongs only to
one subset~$X^i=\{x,y\}$; if the vertex~$y$ also belongs only to
the subset~$X^i$, then the instance of~$\tilde{{\mathcal I}}$ has
no feasible solutions and no further transformations are required;
otherwise assume $X^i=\{x\}$ and label the vertex~$x$ and the
subset~$X^i$ as fathomed.

It is clear that the obtained family of requisitions
is equivalent to the original one and satisfies
conditions~$\mathcal{C}1$~--~$\mathcal{C}3$.
In sequel, without
loss of generality we assume that the family of requisitions
in~$\tilde{{\mathcal I}}$
satisfies~$\mathcal{C}1$~--~$\mathcal{C}3$.

Now let us construct a Turing reduction of
problem~$\tilde{{\mathcal I}}$ to problem~${{\mathcal I}}$.
Suppose there exists a subroutine~${\mathcal{S}}$ for solving
problem~${\mathcal I}$ with a family of requisitions~$\bar{X}^i,\
i=1,\dots,n$. Let us describe an algorithm~${\mathcal{A}}$ for
solving problem~$\tilde{{\mathcal I}}$ with a family of
requisitions~${X}^i,\ i=1,\dots,n$, which applies the
subroutine~$S$ at most four times to supplementary instances
of~${\mathcal I}$, obtained from the original instance by fixing
one of the elements in requisitions~$X^1$ and $X^n$. Note that
such a fixing may violate Condition~$\mathcal{C}3$. If this
happens, the family of requisitions obtained in
algorithm~${\mathcal{A}}$ is transformed into an equivalent one,
complying with conditions~$\mathcal{C}1$~--~$\mathcal{C}3$.
Let us outline the proposed algorithm.\\

{\bf Algorithm~${\mathcal{A}}$}\\
\\
{\bf 1.}~Let $\tilde{f}'$ denote the best found solution to the
instance of~$\tilde{{\mathcal I}}$ and let~$\tilde{\rho}'$ be
value of objective function of this solution. Assign initially~$\tilde{\rho}':=+\infty$.\\
{\bf 2.}~Perform Steps 2.1-2.2 for each vertex~$x\in X^1$:\\
{\bf 2.1.}~Assign $\tilde{X}^1:=\{x\},$ $\tilde{X}^i:=X^i, \
i=2,\dots,n$. Now if $|X^1|=2$, then the family of
requisitions~$\tilde{X}^i,\ i=1,\dots,n$ needs to be transformed
to satisfy Condition~$\mathcal{C}3$. To this end, an index~$j\ne
1$ is found, such that $\tilde{X}^j=\{x,z\}$, and an assignment
$\tilde{X}^j=\{z\}$ is made.
Further perform the similar operations with the vertex~$z$ etc.\\
{\bf 2.2.}~For each vertex $y\in \tilde{X}^n$ perform Steps~2.2.1-2.2.2:\\
{\bf 2.2.1.}~Assign $\bar{X}^n:=\{y\},$ $\bar{X}^i:=\tilde{X}^i, \
i=1,\dots,n-1,$ and if $|\tilde{X}^n|=2$, then transform the
family of requisitions $\bar{X}^i,\ i=1,\dots,n$ so that
Condition~$\mathcal{C}3$ is satisfied, analogously to Step~2.1.\\
{\bf 2.2.2.}~Solve problem~${\mathcal I}$ using
Algorithm~${\mathcal{S}}$. Let $f^*$ be a solution to this
problem. If~$\rho(f^*)+\rho(\bar{X}^n,\bar{X}^1)<\tilde{\rho}'$,
then assign $\tilde{\rho}':=\rho(f^*)+\rho(\bar{X}^n,\bar{X}^1)$
and $\tilde{f}':=f^*$.\\

It is clear that the solution~$\tilde{f}'$ found by
algorithm~${\mathcal{A}}$ will be optimal for
problem~$\tilde{{\mathcal I}}$. Now since $|X^1|\leqslant 2$,
$|X^n|\leqslant 2$, and the transformation of a family of
requisitions takes~$O(n^2)$ time, so the reduction is polynomially
computable. The properties of this reduction imply that
problem~${\mathcal I}$ is strongly NP-hard. $\Box$\\

Therefore the following theorem holds.

\begin{theorem} {\rm \cite{ErKo12}}
The ORP for problem ${1|s_{vu}|C_{\max}}$ is strongly {\rm
NP}-hard.\label{prop2}
\end{theorem}

Although in problem~${\mathcal I}$ we are given a {\em
digraph}~$G$, this problem easily reduces to its modification
where $G$ is an ordinary {\em graph}. This is done by a
substitution of each vertex by three vertices (see
e.g.~\cite{Karp}) and defining an appropriate family of
requisitions~${X^i,\ i=1,\dots,n}$. Therefore the modification of
problem~${\mathcal I}$ on ordinary graphs is also strongly NP-hard
and the next result holds.

\begin{theorem} {\rm \cite{ErKo12}}
The ORP for problem ${1|s_{vu}=s_{uv}|C_{\max}}$ is strongly {\rm
NP}-hard.\label{prop2sym}
\end{theorem}

\subsection{Solving the Optimal Recombination Problem}
\label{subsec:solution} \sloppy
\par

Given an ORP instance of ${1|s_{vu}|C_{\max}}$ problem with parent
solutions~$\pi^1,\pi^2$, one can define an instance of~${\mathcal
I}$ as follows.
\begin{itemize}
\item Let the number of vertices of digraph~$G$ be~$n=k$.
\item Let each job $v_i\in V$, ${i=1,\dots,k}$, be assigned a
vertex~$x_i\in X$ of digraph~$G$.
\item Let the arc weights
be~$\rho(x_i,x_j)=s_{v_i,v_j}$ for all $x_i,x_j\in X$, $i\ne j$.
\item Let the family of requisitions~$X^i$, $i=1,\dots,k$, be such that
$X^i=\{\pi^1_i,\pi_i^2\}$ for those~$i$ where $\pi^1_i \ne
\pi_i^2$ and $X^i=\{\pi^1_i\}$ for the rest of the indices~$i$.
\end{itemize}
In this case, the set of feasible solutions to problem~${\mathcal
I}$ can be mapped to the set of feasible solutions to the ORP for
${1|s_{vu}|C_{\max}}$ by a bijective mapping so that optimal
solutions to problem~${\mathcal I}$ correspond to optimal
solutions to the ORP.

An optimal mapping~$f^*\in F$ for problem~${\mathcal I}$ can be
found in time~$O(2^k)$ by enumeration of all sequences~$\pi$ where
$\pi_i \in X^i$, $i=1,\dots,k$ (feasible as well as infeasible).
An obvious modification of the well-known dynamic programming
algorithm due to M.~Held and R.M.~Karp~\cite{HK} has the same time
complexity. It is possible, however, to build a more efficient
algorithm for solving problem~${\mathcal I}$, using the approach
of A.I.~Serdyukov~\cite{SAI} which was developed for estimation of
cardinality of the set of feasible solutions to
problem~$\tilde{{\mathcal I}}$.

Consider a bipartite graph~$\bar{G}=(X_k,X,\bar{U})$ defined
above. Note that there is a one-to-one correspondence between the
set of feasible solutions~$F$ to problem~${\mathcal I}$ and the
set of perfect matchings~$\mathcal{W}$ in graph~$\bar{G}$.

An edge~$(i,x)\in \bar{U}$ will be called {\em special}, if
$(i,x)$ belongs to all perfect matchings in graph~$\bar{G}$. Let
us also call the vertices of graph~$\bar{G}$ {\em special}, if
they are incident to special edges. A maximal (by inclusion)
bi-connected subgraph~\cite{KLR}
will be called a {\em block}.  Note that in each block~$j$ of
graph~$\bar{G}$ the degree of any vertex equals two,
$j=1,\dots,q(\bar{G})$, where $q(\bar{G})$~denotes the number of
blocks in graph~$\bar{G}$. Then the edges $(i,x)\in \bar{U}$, such
that $|X^i|=1$, are special and belong to none of the blocks,
while the edges $(i,x)\in \bar{U}$, such that $|X^i|=2$, belong to
some blocks. Besides that, each block~${j,\
j=1,\dots,q(\bar{G})}$, of graph~$\bar{G}$ contains exactly two
maximal (edge disjoint) matchings, so it does not contain the
special edges. Hence an edge~$(i,x)\in \bar{U}$ is special iff
$|X^i|=1$, and every perfect matching in~$\bar{G}$ is defined by a
combination of maximal matchings chosen in each of the blocks and
the set of all special edges.

\begin{figure}[!h]
\begin{center}
\vspace{2em}
\includegraphics[height=7.5cm,width=6cm]{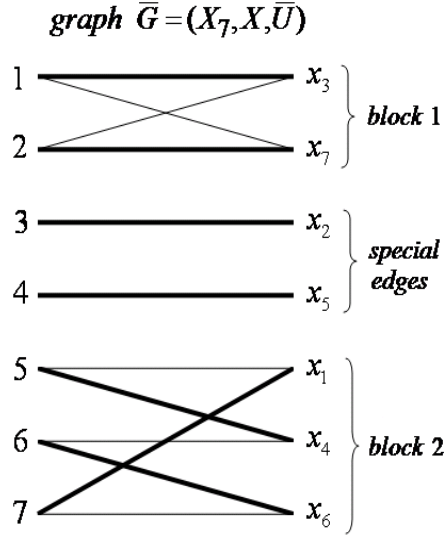}\\
\caption{Example of a graph~$\bar{G}=(X_7,X,\bar{U})$ with two
special edges and two blocks.} \label{fig1JVK}
\end{center}
\end{figure}

As an example consider an instance of~${\mathcal I}$ with $n=k=7$
and the family of requisitions~$X^{1}=\{x_3,x_7\}$,
$X^{2}=\{x_3,x_7\}$, $X^{3}=\{x_2\}$, $X^{4}=\{x_5\}$,
$X^{5}=\{x_1,x_4\}$, $X^{6}=\{x_4,x_6\}$, ${X^{7}=\{x_1,x_6\}}$.
The bipartite graph~$\bar{G}=(X_7,X,\bar{U})$ corresponding to
this problem is presented in Fig.~\ref{fig1JVK}. Here the edges
drawn in bold define one maximal matching of a block, and the rest
of the edges in the block define another one.

The blocks of graph~$\bar{G}$ may be computed in~$O(k)$ time,
e.~g. by means of the "depth first" algorithm~\cite{KLR}. The
special edges and maximal matchings in blocks may be found easily
in $O(k)$~time.

Therefore, the problem~${\mathcal I}$ is solvable by the following
algorithm: Build the bipartite graph~$\bar{G}$, identify the set
of special edges and blocks and find all maximal matchings in
blocks. Enumerate all perfect matchings~$W\in \mathcal{W}$ of
graph~$\bar{G}$ by combining the maximal matchings of blocks and
joining them with special edges. Assign the corresponding
solution~$f\in F$ to each~$W\in \mathcal{W}$ and
compute~$\rho(f)$. As a result one can find~$f^*\in F$, such that
$\rho(f^*)=\min\limits_{f\in F}\rho(f)$.

Note that $|F|=|\mathcal{W}|=2^{q(\bar{G})}$, so the time
complexity of the above algorithm is~$O(k 2^{q(\bar{G})})$, where
${q(\bar{G})\leqslant \lfloor \frac{k}{2}\rfloor}$ and this bound
is tight. Below we propose a modification of this algorithm with
time complexity~$O(q(\bar{G})\cdot 2^{q(\bar{G})})$.

Let us carry out some preliminary computations before enumerating
all possible combinations of maximal matchings in blocks in order
to speed up the evaluation of objective function.
We will call a {\em contact between block~$j$ and block~$j'\ne j$
(or between block~$j$ and a special edge)} the pair of
vertices~$(i,i+1)$ in the left-hand part of graph~$\bar{G}$, such
that one of the vertices belongs to the block~$j$ and the other
one belongs to block~$j'$ (or the special edge). A {\em contact
inside a block} will mean a pair of vertices in the left-hand part
of a block, if their indices differ exactly by one.

For each block~$j,$ ${j=1,\dots,q(\bar{G})}$, let us check the
presence of contacts inside the block~$j$, between the block~$j$
and all special edges, and between the block~$j$ and every other
block. The time complexity of checking for contacts all vertices
in the left-hand part of a block is~$O(k)$.

Consider a block~${j}$. If a contact~$(i,i+1)$ is present inside
this block, then each of the two maximal matchings~$w^{0,j}$ and
$w^{1,j}$ in this block corresponds to an arc of graph~$G$. Also,
if block~$j$ has a contact to a special edge, each of the two
maximal matchings~$w^{0,j}$ and $w^{1,j}$ also corresponds to an
arc of graph~$G$. For each of the matchings~$w^{k,j}, \ k=0,1$,
let the sum of the weights of arcs corresponding to the contacts
inside block~$j$ and the contacts to special edges be denoted
by~$P^{k}_j$.

If block~${j}$ contacts to block~${j', \ j'\ne j}$, then each
combination of the maximal matchings of these blocks corresponds
to an arc of graph~$G$ for any contact~$(i,i+1)$ between the
blocks. If a maximal matching is chosen in each of the blocks, one
can sum up the weights of the arcs in~$G$ that correspond to all
contacts between blocks~${j}$ and ${j'}$. This yields four values
which we denote by $P^{(0,0)}_{j j'}$, $P^{(0,1)}_{j j'}$,
$P^{(1,0)}_{j j'}$ and $P^{(1,1)}_{j j'}$, where the superscripts
identify the matchings chosen in each of the blocks~$j$ and $j'$
accordingly.

The above mentioned sums are computed for each block, so the
overall time complexity of this pre-processing procedure
is~$O(k\cdot q(\bar{G}))$.

Now all possible combinations of the maximal matchings in blocks
may be enumerated using a Grey code (see e.g.~\cite{RND}) so that
the next combination differs from the previous one by altering a
maximal matching only in one of the blocks. Let the binary
vector~$\delta=(\delta_1,\dots,\delta_{q(\bar{G})})$ define
assignments of the maximal matchings in blocks. Namely,
$\delta_j=0$, if the matching~$w^{0,j}$ is chosen in block~$j$;
otherwise (if the matching~$w^{1,j}$ is chosen in block~$j$), we
have~$\delta_j=1$. This way every vector~$\delta$ is bijectively
mapped into a feasible solution~$f_{\delta}$ to problem~${\mathcal
I}$.

In the process of enumeration, a step from the current
vector~$\bar{\delta}$ to the next vector~$\delta$ changes the
maximal matching in one of the blocks~$j$. The new value of
objective function~$\rho(f_{\delta})$ may be computed via the
current value~$\rho(f_{\bar{\delta}})$ by the formula
$\rho(f_{\delta})=\rho(f_{\bar{\delta}})-P^{{\bar
\delta}_j}_j+P^{{\delta}_j}_j-\sum\limits_{j'\in A(j)}P^{({\bar
\delta}_j,{\bar \delta}_{j'})}_{jj'}+\sum\limits_{j'\in
A(j)}P^{({\delta}_j,{\delta}_{j'})}_{jj'}$, where $A(j)$ is the
set of blocks contacting to block~$j$. Obviously, $|A(j)|\leqslant
q(\bar{G})$, so updating the objective function value for the next
solution requires~$O(q(\bar{G}))$ time, and the overall time
complexity of the modified algorithm for solving
Problem~${\mathcal I}$ is~$O(q(\bar{G})\cdot2^{q(\bar{G})})$.

Therefore, the ORP for~$1|s_{vu}|C_{\max}$, as well as
Problem~${\mathcal I}$, is solvable in
$O(q(\bar{G})\cdot2^{q(\bar{G})})$ time. Below it will be shown
that for almost all pairs of parent solutions
${q(\bar{G})}\leqslant 1.1\cdot{\rm ln}(k)$, \ie the cardinality
of the set of feasible solutions in almost all instances of the
ORP for~$1|s_{vu}|C_{\max}$ is at most~$k$ and these instances are
solvable in~$O(k\cdot {\rm ln}(k))$ time.

\begin{definition}{\rm~\cite{SAI}} \label{def:good_graph}
A graph~${\bar{G}=(X_k,X,\bar{U})}$ is called "good" if
${q(\bar{G})\leqslant 1.1\cdot{\rm ln}(k)}$; otherwise it is
called "bad".
\end{definition}

\begin{definition} \label{def:good_rec}
A pair of parent solutions $\{\pi^1, \pi^2\}$ is called "good" if
the graph~$\bar{G}=(X_k,X,\bar{U})$ corresponding to these parent
solutions is "good"; otherwise the pair~$\{\pi^1, \pi^2\}$ is
called "bad".
\end{definition}

Note that instead of constant~$1.1$ in
Definition~\ref{def:good_graph} one may choose any other constant
equal to~$1+\varepsilon,$ where $\varepsilon\in (0,{\rm
log}_2(e)-1]$. Given such a constant, the ORP has at most~$k$
feasible solutions and it is solvable
in~$O(k {\rm ln}(k))$ time.\\

The following notation will be used below:
\begin{itemize}
\item Let~$\bar{\Im}_k$ be the set of "good" graphs and
let~$\tilde{\Im}_k$ denote the set of "bad" graphs.

\item Let $\bar{\Re}_k$ be the set of "good"
pairs of parent solutions and let~$\tilde{\Re}_k$ be the set of
"bad" pairs of parent solutions.

\item Denote ${\Im}_k=\bar{\Im}_k\cup \tilde{\Im}_k$,
${\Re}_k=\bar{\Re}_k\cup \tilde{\Re}_k$.

\item Let ${S}_l$~be the set of permutations of the set~$\{1,\dots,l\}$, 
which do not contain the cycles of length~1.

\item Let $\bar{S}_l$~denote the set of permutations from~$S_l$, where the
number of cycles is at most~${1.1\cdot{\rm ln}(l)}$.

\item Denote $\tilde{S}_l={S}_l\backslash \bar{S}_l$.
\end{itemize}

The results of A.I.~Serdyukov from~\cite{SAI} imply

\begin{proposition}
$|\tilde{S}_l|/|\bar{S}_l| \longrightarrow 0$ as $l\to
\infty$.\label{th3}
\end{proposition}

The next theorem is proved by the means of Proposition~\ref{th3}.
\begin{theorem} {\rm \cite{ErKo12}}
$|\bar{\Re}_k|/|{\Re}_k| \longrightarrow 1$ as $k\to
\infty$.\label{th:good_rec}
\end{theorem}

{\bf Proof.} The proof consists of two stages: first we estimate
the numbers of "good" and "bad" graphs, and after that we estimate
the numbers of "good" and "bad" pairs of parent solutions.

The values~$|\bar{\Im}_k|$ and $|\tilde{\Im}_k|$ may be bounded
using the approach from~\cite{SAI}. To this end assign any
permutation~$\sigma \in {S}_l$, $l\leqslant k$, a set of
bi-partite graphs~${{\Im}_k(\sigma)\subset {\Im}_k}$ as follows.
First of all let us assign an arbitrary set of $k-l$ edges to be
special. The non-special vertices $\{i_1,i_2,\dots,i_l\}\subset
X_k$ of the left-hand part, where $i_j<i_{j+1},\ j=1,\dots,l-1$,
are now partitioned into $\xi(\sigma)$ blocks, where $\xi(\sigma)$
is the number of cycles in permutation~$\sigma$. Every cycle
$(t_1,t_2,\dots,t_r)$ in permutation~$\sigma$ corresponds to some
sequence of vertices with indices
$\{i_{t_1},i_{t_2},\dots,i_{t_r}\}$ 
belonging to the block associated with this cycle. Finally, it is
ensured that for each pair of vertices $\{i_{t_j},\
i_{t_{j+1}}\}$, $j=1,\dots,r-1$, as well as for the pair
$\{i_{t_{r}}, \ i_{t_1}\},$ there exists a vertex in the
right-hand part~$X$ adjacent to both vertices of the pair.

\begin{figure}[!h]
\begin{center}
\vspace{2em}
\includegraphics[height=7.5cm,width=13cm]{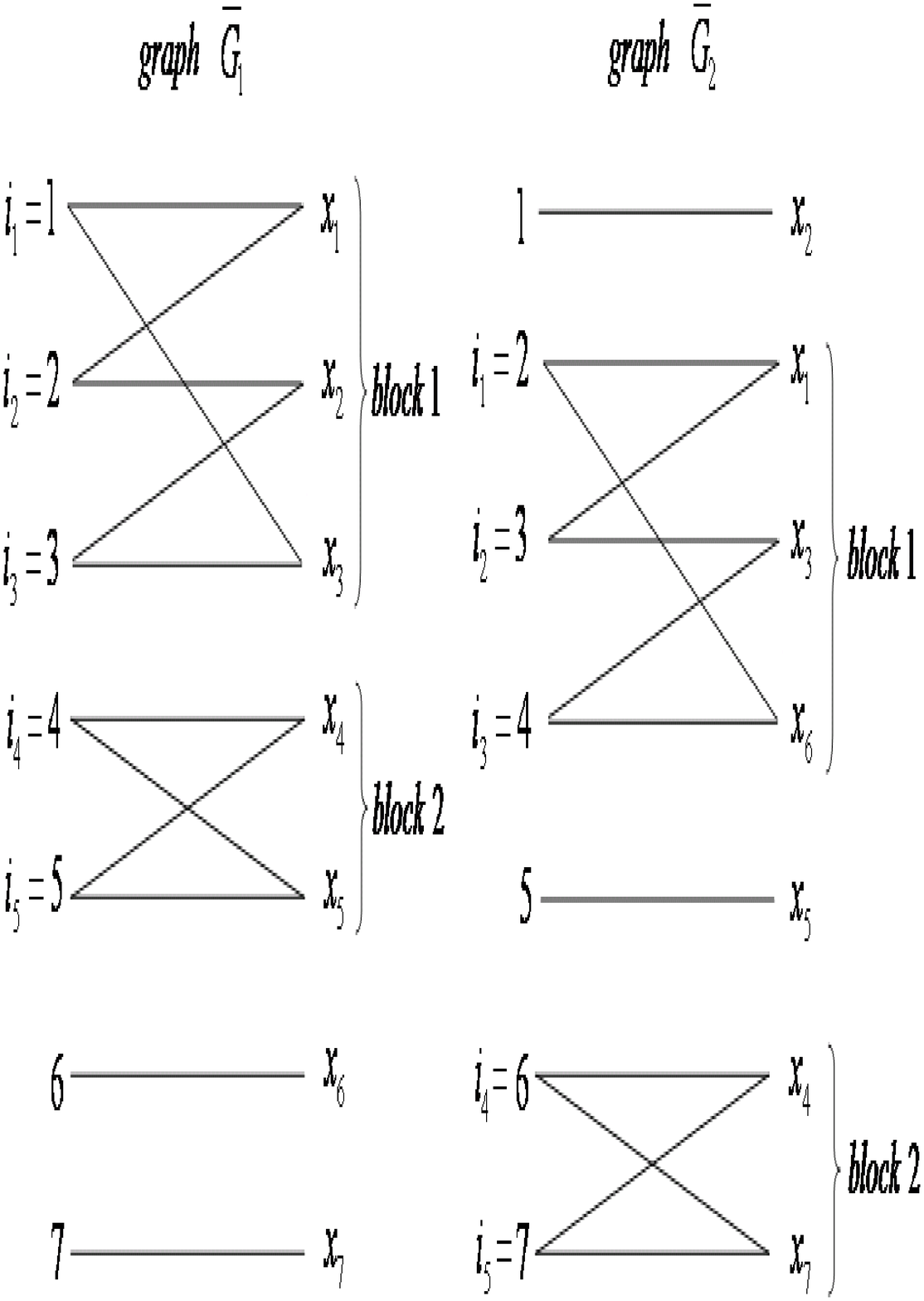}\\
\caption{Examples of graphs from class~${{\Im}_7(\sigma)}$, where
$\sigma={1\ 2\ 3\ 4\ 5 \choose 2\ 3\ 1\ 5\ 4} \in S_5$.}
\label{fig2JVK}
\end{center}
\end{figure}

Consider a permutation~$\sigma={1\ 2\ 3\ 4\ 5 \choose 2\ 3\ 1\ 5\
4} \in S_5$ with cycles $c_1=(1,2,3)$ and $c_2=(4,5)$. Two
examples of graphs from class~${{\Im}_7(\sigma)}$ are given in
Fig.~\ref{fig2JVK}. Here block~$j$ corresponds to cycle~$c_j$,
$j=1,2$.

There are~$k!$ ways to associate vertices of the left-hand pert to
vertices of the right-hand part, therefore the number of different
graphs from class~${\Im}_k(\sigma)$, $\sigma \in {S}_l$,
$l\leqslant k$, is
$|{\Im}_k(\sigma)|=C_k^l\frac{k!}{2^{\xi_1(\sigma)}}$, where
$\xi_1(\sigma)$ is the number of cycles of length two in
permutation~$\sigma$. Division by~$2^{\xi_1(\sigma)}$ here is due
to the fact that for each block that corresponds to a cycle of
length two in~$\sigma$, there are two equivalent ways to number
the vertices in its right-hand part.

Let~${\sigma=c_1c_2\dots c_{\xi(\sigma)}}$ be a permutation from
set~${S}_l$, represented by cycles~$c_i$,
${i=1,\dots,\xi(\sigma)},$ and let $c_j$ be an arbitrary cycle of
permutation~$\sigma$ of length at least three, $1\leqslant j
\leqslant \xi(\sigma)$. Permutation~$\sigma$ may be transformed
into permutation~$\sigma^1$,

\begin{equation}
\sigma^1=c_1c_2\dots c_{j-1} c_j^{-1}c_{j+1}\dots
c_{\xi(\sigma),}\label{cicle}
\end{equation}
by reversing the cycle~$c_j$. Clearly, permutation~$\sigma^1$
induces the same subset of graphs in class~${\Im}_k$ as the
permutation~$\sigma$ does. Thus any two permutations~$\sigma^1$
and $\sigma^2$ from set~${S}_l$, $l\leqslant k$, induce the same
subset of graphs in~${\Im}_k$, if one of these permutations may be
obtained from the other one by several transformations of the
form~(\ref{cicle}). Otherwise the two induced subsets of graphs do
not intersect. Besides  that ${\Im}_k(\sigma^1)\cap
{\Im}_k(\sigma^2)=\emptyset$ if $\sigma^1\in {S}_{l_{1}}$,
$\sigma^2\in {S}_{l_{2}}$, $l_1\ne l_2$.

On one hand, if $\sigma\in \bar{S}_l$, $l\leqslant k$,  then
${\Im}_k(\sigma)\subseteq {\bar{\Im}}_k$. On the other hand, if
$\sigma\in \tilde{S}_l$, $l< k$, then either
${\Im}_k(\sigma)\subseteq {\bar{\Im}}_k$ or, alternatively,
${\Im}_k(\sigma)\subseteq {\tilde{\Im}}_k$ may hold. Therefore,

\begin{equation}
|\bar{\Im}_k|\geqslant\sum_{l=2}^k\sum_{\sigma \in
\bar{S}_l}C_k^l\frac{k!}{2^{\xi_1(\sigma)}2^{\xi(\sigma)-\xi_1(\sigma)}}=\sum_{l=2}^k\sum_{\sigma
\in \bar{S}_l}C_k^l\frac{k!}{2^{\xi(\sigma)}}, \label{eq1}
\end{equation}

\begin{equation}
|\tilde{\Im}_k|\leqslant \sum_{l=\lfloor 1.1\cdot{\rm
ln}(k)\rfloor}^k\sum_{\sigma \in
\tilde{S}_l}C_k^l\frac{k!}{2^{\xi_1(\sigma)}2^{\xi(\sigma)-\xi_1(\sigma)}}=\sum_{l=\lfloor
1.1\cdot{\rm ln}(k)\rfloor}^k\sum_{\sigma
\in \tilde{S}_l}C_k^l\frac{k!}{2^{\xi(\sigma)}}.\\
\label{eq2}
\end{equation}

Now let us estimate the cardinality of sets $\bar{\Re}_k$ and
$\tilde{\Re}_k$ to complete the proof. Recall that every
graph~${\bar{G}\in {\Im}_k(\sigma),}$ $\sigma\in {S}_l,$
$l\leqslant k$ has~$\xi(\sigma)$ blocks. The set of edges of any
block~$j$, ${j=1,\dots,\xi(\sigma)}$, is partitioned into the
maximal matchings denoted by
$w^j=\{(i_1,x^{i_1}),(i_2,x^{i_2}),\dots,(i_{m_j},x^{i_{m_j}})\}$
and
$\bar{w}^j=\{(i_1,\bar{x}^{i_1}),(i_2,\bar{x}^{i_2}),\dots,(i_{m_j},\bar{x}^{i_{m_j}})\}$.
Then in any instance of the ORP for problem~$1|s_{vu}|C_{\max}$,
that induces the graph~$\bar{G}$, either $\pi^1_{i_m}=x^{i_m}$,
$\pi^2_{i_m}=\bar{x}^{i_m}$, $m=1,\dots,m_j$, or
$\pi^1_{i_m}=\bar{x}^{i_m}$, $\pi^2_{i_m}={x}^{i_m}$,
$m=1,\dots,m_j$, for all $j=1,\dots,\xi(\sigma)$. Consequently
every bipartite graph from class~${\Im}_k(\sigma)$ corresponds
to~$2^{\xi(\sigma)}$ pairs of parent solutions (where pairs
$\pi^1=a$, $\pi^2=b$ and $\pi^1=b$, $\pi^2=a$ are assumed to be
different), then in view of~(\ref{eq1})~and~(\ref{eq2}) we have:

\begin{equation}
|\bar{\Re}_k|\geqslant\sum_{l=2}^k\sum_{\sigma \in
\bar{S}_l}C_k^l\frac{k!}{2^{\xi(\sigma)}}2^{\xi(\sigma)}\geqslant
\sum_{l=\lfloor 1.1\cdot{\rm ln}(k)\rfloor}^k |\bar{S}_{l}|
C_k^lk!, \label{eq3}
\end{equation}

\begin{equation}
|\tilde{\Re}_k|\leqslant \sum_{l=\lfloor 1.1\cdot{\rm
ln}(k)\rfloor}^k\sum_{\sigma \in
\tilde{S}_l}C_k^l\frac{k!}{2^{\xi(\sigma)}}2^{\xi(\sigma)}=
\sum_{l=\lfloor 1.1\cdot{\rm ln}(k)\rfloor}^k |\tilde{S}_{l}|
C_k^lk!. \label{eq4}
\end{equation}

Now assuming $\psi(k)=\max\limits_{l=\lfloor 1.1\cdot{\rm
ln}(k)\rfloor,\dots,k}|\tilde{S}_{l}|/|\bar{S}_{l}|$ and taking
into account~(\ref{eq3}),~(\ref{eq4}) and Proposition~\ref{th3},
we obtain

\begin{equation}
|\tilde{\Re}_k|/|\bar{\Re}_k|\leqslant \psi(k) \to 0 \mbox{ as }
k\to \infty. \label{eq5}
\end{equation}
Finally, the statement of the theorem follows from~(\ref{eq5}).
$\Box$\bigskip

Note that the algorithm proposed for solving the ORP
for~$1|s_{vu}|C_{\max}$ may be generalized to solve the ORPs for
other problems with similar solutions encoding (examples of such
problems may be found in~\cite{HGE,TKSH,YI96}). The time
complexity of the algorithm in these cases would depend on the
time required to evaluate an objective function.

Theorems~\ref{prop2} and \ref{prop2sym} imply NP-hardness of the
ORPs for a family of more general scheduling problems, where the
number of machines may be greater than~1 and each job may be
performed in several modes, using one or more machines, see
e.g.~\cite{ErKo11}.

\section{Conclusion \label{Conclusion}}

We have shown that optimal recombination may be efficiently
carried out for many important NP-hard optimization problems. The
well-known reductions between the NP~optimization problems turned
out to be useful in development of polynomial-time optimal
recombination procedures. We have observed that the choice of
solutions encoding has a significant influence upon the complexity
of the optimal recombination problems and introduction of
additional variables can sometimes simplify the task (compare
Corollary~\ref{setpart_splp} and Proposition~\ref{SPLP_NP}). The
question of practical utility of such simplifications remains
open, since the additional redundancy in representation increases
the number of constraints in the ORP. This trade-off may be
studied in further research.

Another open question is related to the trade-off between the
complexity of optimal recombination and its impact on the
efficiency of an evolutionary algorithm (e.g. in terms of
optimization time).
The
theoretical methods proposed in~\cite{dhk12} and~\cite{RRS98} may
be helpful in runtime analysis of GAs with optimal recombination.

All of the polynomially solvable cases of the optimal
recombination problems considered above rely upon the efficient
deterministic algorithms for the Max-Flow/Min-Cut Problem (or the
Maximum Matching Problem in the unweighted case). However, the
crossover operator was initially introduced as a randomized
operator in genetic algorithms~\cite{Holl75}. As a compromise
approach one can solve the optimal recombination problem
approximately or solve it optimally but not in all occasions.
Examples of the genetic algorithms using this approach may be
found in~\cite{BDE,DEG10,ErOR,ErKo11}.

The obtained results indicate that optimal recombination for many
NP-hard optimization problems is also NP-hard. It is natural to
expect, however, that the ORP instances emerging in a GA would
often have much smaller dimensions, compared to the original
problem. The average dimensions of the ORP might decrease in
process of GA execution, as the individuals gain more common
genes. In such situations even the NP-hard ORP may turn out to be
solvable in practice by the exact methods, see
e.g.~\cite{AOT00,DEG10,ErKo11}.


In this paper, we did not discuss the population management
strategies of the GAs with optimal recombination. Due to fast
localization of the search process such GAs, it is often important
to provide a sufficiently large initial population and employ some
mechanism for adaptation of the mutation strength. Interesting
techniques that maintain the diversity of population by
constructing the second offspring, as different from the optimal
offspring as possible, can be found in~\cite{AOT97}
and~\cite{BN98}. It is likely that the general schemes of the
evolutionary algorithms and the procedures of parameter adaptation
require some revision when the optimal recombination is used (see
e.g.~\cite{DEG10,YI96}).

\section{Acknowledgements} Partially supported by Russian Foundation
for Basic Research grants 12-01-00122 and 13-01-00862 and by
Presidium SB RAS (project 7B).

\end{document}